\documentclass[lettersize,journal]{IEEEtran}
\usepackage{amsmath,amsfonts}
\usepackage{algorithmic}
\usepackage{algorithm}
\usepackage{array}
\usepackage[caption=false,font=small,labelfont=sf,textfont=sf]{subfig}
\usepackage{textcomp}
\usepackage{stfloats}
\usepackage{url}
\usepackage{verbatim}
\usepackage{graphicx}
\usepackage{lipsum}

\usepackage{cite}
\usepackage{amsthm}
\usepackage{makecell}
\usepackage{multirow}
\usepackage{multicol}
\usepackage{booktabs}
\newcommand\SAGNAS{SA-GNAS}
\newtheorem{myDef}{Definition}
\newtheorem{prop}{Proposition}
\usepackage{graphicx, caption} 

\begin{document}

\title{SA-GNAS: Seed Architecture Expansion for Efficient Large-scale Graph Neural \\ Architecture Search}

\author{Guanghui~Zhu,~\IEEEmembership{Member,~IEEE,}
	    Zipeng~Ji, Jingyan~Chen,
     Limin~Wang,~\IEEEmembership{Member,~IEEE} \\ Chunfeng~Yuan, and~Yihua~Huang
\thanks{Guanghui~Zhu, Zipeng~Ji, Jingyan~Chen, Limin~Wang, Chunfeng~Yuan, and Yihua~Huang are with State Key Laboratory for Novel Software Technology, Nanjing University, Nanjing, China. E-mail: zgh@nju.edu.cn, \{jizipeng, chenjy\}@smail.nju.edu.cn, lmwang.nju@gmail.com, \{cfyuan, yhuang\}@nju.edu.cn. Guanghui Zhu is the corresponding author.}
\thanks{Manuscript received April 19, 2021; revised August 16, 2021.}}

\markboth{Journal of \LaTeX\ Class Files,~Vol.~14, No.~8, August~2021}%
{Shell \MakeLowercase{\textit{et al.}}: A Sample Article Using IEEEtran.cls for IEEE Journals}

\IEEEpubid{0000--0000/00\$00.00~\copyright~2021 IEEE}

\maketitle

\begin{abstract}
GNAS~(Graph Neural Architecture Search) has demonstrated great effectiveness in automatically designing the optimal graph neural architectures for multiple downstream tasks, such as node classification and link prediction.
However, most existing GNAS methods cannot efficiently handle large-scale graphs containing more than million-scale nodes and edges  due to the expensive computational and memory overhead. 
%
To scale GNAS on large graphs while achieving better performance, we propose \SAGNAS, a novel framework based on \emph{seed architecture
expansion} for efficient large-scale GNAS.
Similar to the \emph{cell expansion} in biotechnology, we first construct a seed architecture and then expand the seed architecture iteratively.
Specifically, we first propose a performance ranking consistency-based seed architecture selection method, which selects the architecture searched on the subgraph that best matches the original large-scale graph.
Then, we propose an entropy minimization-based seed architecture expansion method to further improve the performance of the seed architecture.
Extensive experimental results on five large-scale graphs demonstrate that the proposed \SAGNAS{} outperforms human-designed state-of-the-art GNN architectures and existing graph NAS methods.
Moreover, \SAGNAS{} can significantly reduce the search time, showing better search efficiency.
For the largest graph with billion edges, \SAGNAS{} can achieve $2.8\times$ speedup compared to the SOTA large-scale GNAS method GAUSS.
Additionally, since \SAGNAS{} is inherently parallelized, the search efficiency can be further improved with more GPUs.
\SAGNAS{} is available at \url{ https://github.com/PasaLab/SAGNAS}.

%
\end{abstract}


\section{Introduction}
\IEEEPARstart{I}{n} recent years, Graph Neural Networks (GNNs)~\cite{hamilton2018representation,kipf2017gcn,velivckovic2018gat,hamilton2017sage} 
have been demonstrated to be a powerful tool for learning from graph-structured data. 
Due to  the great success in diverse fields (e.g., social network analysis~\cite{pmlr-v162-wang22am}, traffic prediction~\cite{graphtraffic}, combinatorial optimization~\cite{graphcombinatorial}, molecular property prediction~\cite{graphmecular}, and recommendation systems~\cite{graphrecomm}), GNNs have gained increasing attention from researchers. 
However, although many GNN models have been proposed, there is no single GNN architecture that can achieve the best performance on different graph datasets or tasks. 
Given a specific graph dataset, designing effective GNN architectures often requires extensive prior knowledge and trial-and-error~\cite{gao2022hgnn+, tiezzi2021deep, zhang2022unsupervised}. 
This challenge has spurred the emergence of Graph Network Architecture Search (GNAS)~\cite{GNAS-Survey, gao2020graphnas, gasso, autognn, sane, psp}, which aims to find optimal GNN architectures automatically.

Existing reinforcement learning-based and differentiable GNAS methods have demonstrated excellent learning performance on relatively small-scale graphs, typically containing up to million nodes and edges~\cite{gauss}.
%
%
However, in real-world scenarios, many graphs are extensively large-scale with billions of nodes and edges~\cite{hu2021open}. 
Training a single GNN model on such large-scale graphs incurs substantial time and memory costs~\cite{Jia2020}, which leads to much larger overhead for GNAS due to that a large number of candidate architectures need to be trained and evaluated during the search process.
Consequently, most GNAS methods fail to handle large-scale graphs. 
%
%
%
To make GNN architecture search on large-scale graphs feasible, many efficient search methods have been proposed.
%
%

%
Inspired by existing scalable GNNs models that follow the sampling-then-training paradigm, searching GNN architectures on subgraphs rather than the original large-scale graph presents an effective solution to the scalability issue, where sampling can reduce the training time and memory costs by selecting partial nodes and edges from the entire graph~\cite{gcn-cluster, graphsaint, LayerNeighborSampling}.
%
%
However, such sampling-based GNAS methods (e.g., EGAN~\cite{EGAN}) for large-scale graphs face the performance consistency challenges~\cite{gauss}.
The searched architecture on sampled subgraphs cannot stabilize the performance, which may lead to a performance collapse after transferring to the original large-scale graph. 
%
GAUSS~\cite{gauss}, the first GNAS method for large-scale graphs with more than one billion edges, proposed a joint sampling strategy between
architectures and graphs to mitigate the consistency collapse issue to some extent.
Nevertheless, the reinforcement learning-based search strategy in GAUSS requires a large number of architecture evaluations, leading to relatively low search efficiency (i.e., GPU day-level search overhead on graphs with billion edges).

\IEEEpubidadjcol
To ensure the performance consistency between sampled subgraphs and the entire graph and further improve the search efficiency,  we propose a novel framework based on \textit{seed architecture expansion} for efficient large-scale GNAS (\SAGNAS{} for short). 
Unlike existing GNAS methods that directly search for the final architecture, we first search the seed architecture and then expand the seed architecture iteratively.
The overall search process of \SAGNAS{} is similar to the \textit{cell expansion} in biotechnology.

Specifically, we first propose a \emph{performance ranking consistency}-based seed architecture selection method.
Multiple different subgraphs are sampled from the original large-scale graph, and we perform differentiable architecture search independently for each subgraph within the cell-based micro search space, which allows the architecture search process for multiple subgraphs to be inherently parallel.
To measure the similarity between the subgraph and the original graph, we design a performance-driven matching strategy. 
By evaluating the performance of each searched architectures on all subgraphs and the original graph, we construct corresponding \emph{performance sequences} for each subgraph as well as the original graph.
Next, we employ weighted Kendall $\tau$ coefficient to measure the ranking consistency between performance sequences of each subgraph and the original graph.
A higher ranking consistency indicates that the subgraph more accurately represents the characteristics of the original graph. 
Therefore, we select the architecture searched on the most consistent subgraph as the seed architecture.

To further improve the performance of the seed architecture, we propose an \emph{entropy minimization-based} seed architecture expansion method. 
For the directed acyclic graph (DAG) of the cell-based search space, the edge indicates the message aggregation operation in GNNs.
The \emph{edge entropy} is defined as the uncertainty of the searched operation.
The sum of the entropy of all input edges reflects the entropy of the node, which serves as a measure of the stability of the messages it receives.
During each iteration of seed architecture expansion, we select the node with the largest entropy for splitting and subsequently search the local architecture associated with the split node and the newly generated node, while keeping the rest of the architecture unchanged.
To accommodate the increasing complexity of the architecture, we perform \emph{collaborative expansion} of the subgraph and the architecture in each iteration.
Additionally, we introduce a regularization term designed to constrain the overall entropy of the architecture throughout the localized differential architecture search process.

In summary, the contributions of this paper are as follows:

\begin{itemize}
    \item We propose a novel framework based on \emph{seed architecture expansion} for efficient large-scale graph neural architecture search. A highly promising seed architecture is initially identified and then progressively expanded.
    \item We propose a \emph{performance ranking consistency-based} seed architecture selection method. Weighted kendall $\tau$ coefficient is utilized to measure the ranking consistency between the performance sequences of each subgraph the original graph. The architecture searched on the subgraph that most closely aligns with the original large-scale graph is selected as the seed architecture. 
    \item 
    To further improve the performance consistency, we propose an \emph{entropy minimization-based} seed architecture expansion method by collaboratively increasing subgraph size and architecture complexity in each iteration.
    \item Extensive experimental results on five large-scale graphs demonstrate that the proposed \SAGNAS{} outperforms the SOTA human-crafted GNN architecture and existing GNAS methods in terms of both effectiveness and efficiency. For the large-scale graph with billion edges, the search overhead of \SAGNAS{} can be reduced to approximately 8 GPU hours using a single GPU, and to 2 GPU hours through parallel acceleration with four GPUs.
\end{itemize}

\section{Related Work}
\subsection{Neural Architecture Search}
\label{sec2.1 NAS}
Neural Architecture Search (NAS)~\cite{nas2019jmlr, he2021automl, Ren2021Survey, chen2023understanding, zheng2021migo,zhang2020you,chen2023mngnas,wang2021sample} has developed as a well-formed realm to automatically discover architectures for various deep learning tasks, such as image processing~\cite{darts,nasfpn,autodeeplab,tian2021alphagan}, recommendation system~\cite{nas-ctr, autogsr}. 
%
Given a defined search space $\mathcal{F}$, the training set $D_{\text{train}}$, the validation set $D_{\text{val}}$, and the evaluation metric $M$, we aim to find the optimal architecture $f^{*} \in \mathcal{F}$ that achieves the best metric on the validation set:
\begin{equation}
    f^{*} = \mathop{\arg \max}_{f \in \mathcal{F}} M \left( f \left( \theta^{*}\right), D_{\text{val}} \right),
\end{equation}
where $\theta^{*} = \mathop{\arg \min}_{\theta} L(f(\theta), D_{\text{train}})$ denotes the parameter learned from architecture $f$ by minimizing the loss function $L$ over the training set.
Early efforts employed reinforcement learning~\cite{zoph2017neural, pham2018efficient, NASNet} and evolutionary algorithm~\cite{hierarchical,real2019regularized} to explore the potential architectures.
To enhance the search efficiency, differentiable search methods (DARTS) have been proposed~\cite{darts, yao2020efficient, wang2021rethinking, jiang2024operation,yu2022cyclic,yan2021zeronas}.

DARTS designs a cell-based micro search space, where each cell is represented as a directed acyclic graph. 
%
Each node is a latent representation and each directed edge is associated with an operation $o \in \mathcal{O}$ (e.g. convolution) selected from the candidate operation set $\mathcal{O}$.
DARTS relaxes the categorical choice of a particular operation to a softmax over all possible operations, 
%
%
and utilizes the gradient-based method to jointly optimize architecture parameters and model weights.
%


\subsection{Graph Neural Architecture Search}
Inspired by NAS, a variety of graph neural architecture search methods have been proposed~\cite{GNAS-Survey}. 
Similar to NAS methods, the search strategies on graphs can be categorized into into three typical types: reinforcement learning-based~\cite{gao2020graphnas, autognn,HighPG}, evolutionary algorithm-based~\cite{GeneticGNN, autoattend}, and gradient-based~\cite{zhao2020simplifying,sane,dss} methods. 

GraphNAS~\cite{gao2020graphnas} is the pioneering work that applies reinforcement learning into GNAS.
Meta-GNAS\cite{Meta-GNAS} uses meta-reinforcement learning to improve the search efficiency on multiple tasks. 
Auto-GNN~\cite{autognn} designs a more comprehensive search space and utilizes a reinforced conservative controller to build architectures. 
Additionally, HGNAS~\cite{HGNAS} leverages reinforcement learning to heterogeneous GNNs.

The evolutionary strategy in NAS has also been widely adopted in GNAS. 
AutoGraph~\cite{AutoGraph} employs an evolutionary algorithm to automatically evolve deep GNN models. 
GeneticGNN~\cite{GeneticGNN} updates the GNN architecture and hyperparameters through an evolutionary process, where each network structure is represented by a fixed-length binary string. 
GraphPAS~\cite{GraphPAS} proposes a parallel GNAS framework by designing a sharing-based evolutionary learning approach.

Differentiable NAS methods such as DARTS~\cite{darts} and SGAS~\cite{sgas} have gained great attention in recent years. These methods can significantly improve the efficiency of NAS through parameter-sharing and continuous relaxation. 
%
AutoSTG~\cite{AutoSTG}, SANE~\cite{sane}, and DSS~\cite{dss} adopt the differentiable gradient-based GNAS method to search GNN architectures from the search space. 
Furthermore, NAC~\cite{nac} implements a free-training GNAS method that leverages a sparse coding objective across multiple GNN benchmark datasets.

\subsection{GNAS on Large-Scale Graphs}
Most existing GNAS works conduct their search processes on relatively small-scale graphs, typically consisting of millions of nodes and edges.
However, significant challenges such as neighbor explosion and computational burden arise when dealing with large-scale graphs with billions of nodes and edges~\cite{wang2022automated}. 
Recently, several studies have focused on enhancing the performance of GNAS for large-scale graphs.

GASSO~\cite{gasso} employs graph structure learning as a de-noising process and jointly optimizes neural architecture and graph structure. 
It extends its experiments to three larger graph benchmarks: Physics, Cora-Full, and ogbn-arxiv. 
PaSca~\cite{pasca} introduces a scalable GNAS framework with a novel search space designed for web-scale graphs.
EGAN~\cite{sasaki2023efficient} adapts to moderate-scale graphs by sampling a small subgraph as a proxy, which is then directly applied to large-scale graphs. 
GAUSS~\cite{gauss} achieves joint architecture and graph sampling sampling via a GNN architecture peer learning strategy on the sampled subgraphs. 
HGNAS++~\cite{HGNAS}, building on HGNAS, investigates GNAS on large-scale heterogeneous graphs.

\section{THE PROPOSED Methodology}

\subsection{Search Space Design}
\label{ss:space}
The search space is critical for the effectiveness of graph NAS~\cite{graphpnas,GNAS-Survey}. 
%
Designing GNN architectures within the cell-based search spaces~\cite{sgas,psp} has demonstrated impressive performance and efficiency compared to other GNAS strategies. 
In contrast to the macro search space, where the size increases exponentially with the number of layers, the micro search space consists of multiple cells with identical architectures.
Instead of searching the entire graph neural
network directly, we focus on optimizing the cell architecture.
Inspired by DARTS~\cite{darts} and previous cell-based GNAS approaches~\cite{psp,sgas}, we propose a concise cell-based graph search space for large-scale GNAS.

\begin{figure}[htpb]
    \centering
    \includegraphics[width=1\linewidth]{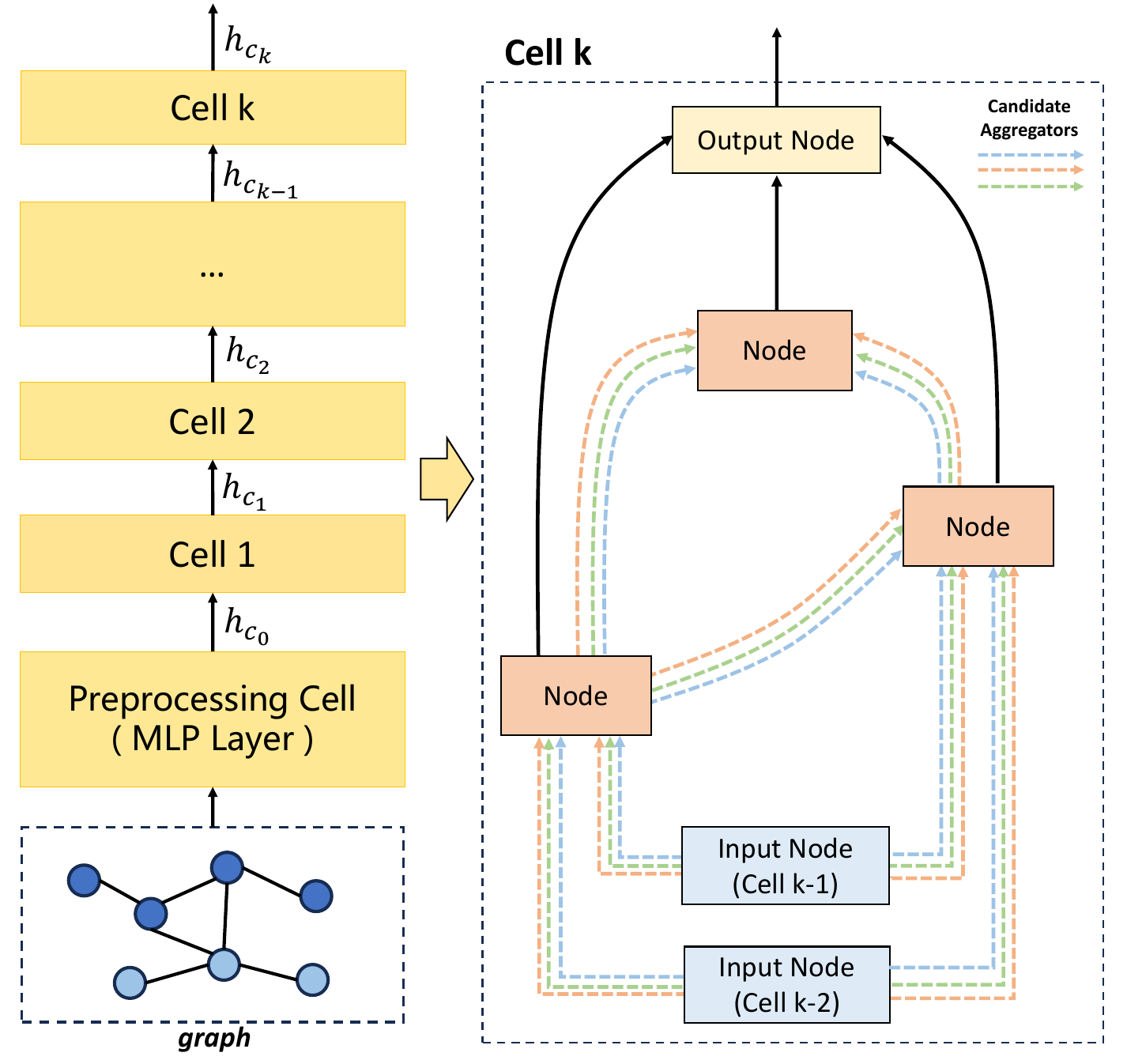}
    \caption{Left: the cell-based search space design. Right: the cell architecture. Different edge colors represent various candidate aggregation operations. The search objective is to identify the optimal aggregation operation on each edge.}
    \label{fig:space-structure}
\end{figure}

As illustrated in Figure~\ref{fig:space-structure}, the architecture of the cell can be represented as a directed acyclic graph, which contains two input nodes, $P$ intermediate nodes, and one output node.
%
For cell $k$, the two input nodes correspond to the outputs of the previous two cells (i.e., cell $k-1$ and cell $k-2$).
Each intermediate node receives the outputs of the previous nodes as its inputs.
The final node concatenates the outputs of all intermediate nodes to produce the overall output. 
%
%
%

Formally, let $N^{(i)}, 0 \le i \le P + 2$ define the $i$-th node. The first two nodes $N^{(0)}$, $N^{(1)}$ are input nodes and the last node $N^{(P + 2)}$ is the output node. The intermediate nodes $\{N^{(i)} \mid 2 \le i \le P + 1\}$ are wired with each other.
For intermediate node $N^{(i)}$ in the cell, the corresponding output $h_{i}$ is as follows:
\begin{equation}
    h_{i} = \sum_{j=0}^{i-1}o(h_{j}), o \in \mathcal{O}, 2 \le i \le P + 1.
\end{equation}
The input edge of $N^{(i)}$ represents the message aggregation operation $o$ in GNN. 
$\mathcal{O}$ denotes the candidate aggregation operations.
The final output of cell $k$ is denoted as:
\begin{equation}
    h_{c_k} = concat(h_{2}, \cdots, h_{{P+1}}). 
\end{equation}

\begin{table}[t] 
\caption{The search space in the cell.}
\resizebox{\linewidth}{!}{
\begin{tabular}{@{}ll|l@{}}
\toprule
\multicolumn{2}{l|}{\textbf{Search Space}} &
  \textbf{Candidate Operations} \\ \midrule
\multicolumn{1}{l|}{\makecell{Message Aggregator \\ in GNN}} &
  parametric &
  \begin{tabular}[c]{@{}l@{}}GCN~\cite{kipf2017gcn}, \\ GAT\_\{1,4,8\}~\cite{velivckovic2018gat}, \\ GAT-COS/LINEAR/GEN-LINEAR~\cite{velivckovic2018gat}, \\ GIN~\cite{pmlr-v162-wang22am}, \\ SAGE-MAX/SUM/MEAN~\cite{hamilton2017sage}, \\ ARMA~\cite{bianchi2021graph},\\ APPNP~\cite{klicpera2018predict} \end{tabular} \\ \cmidrule(l){2-3} 
\multicolumn{1}{l|}{} &
  non-parametric &
  \textit{Zero, Skip\_Connect} \\ \bottomrule
\end{tabular}
}
\label{tbl:searchspace}
\vspace{-1ex}
\end{table}
Table~\ref{tbl:searchspace} shows the search space of the cell architecture. 
To minimize the number of candidate operations, we leverage the design principles of existing well-performing GNNs. 
For instance, GAT\_4 represents the combinations of the sampling function, attention computation function, and aggregation function used in~\cite{velivckovic2018gat} with 4 attention heads, while SAGE-SUM refers to GraphSage~\cite{hamilton2017sage} with the sum aggregator.
The existing hand-crafted combinations in the GNN literature have proven to be effective.
Furthermore, the search space can be further reduced since no combinatorial enumeration is required.
%
%
Specifically, as depicted in Table~\ref{tbl:searchspace}, we employ 13 parametric and 2 non-parametric candidate operations.   
Similar to SGAS~\cite{sgas}, \textit{Skip\_Connect} indicates residual connection, and \textit{Zero} means there is no connection between two nodes.

 \begin{figure*}[t]
\centering
\includegraphics[width=1\textwidth]{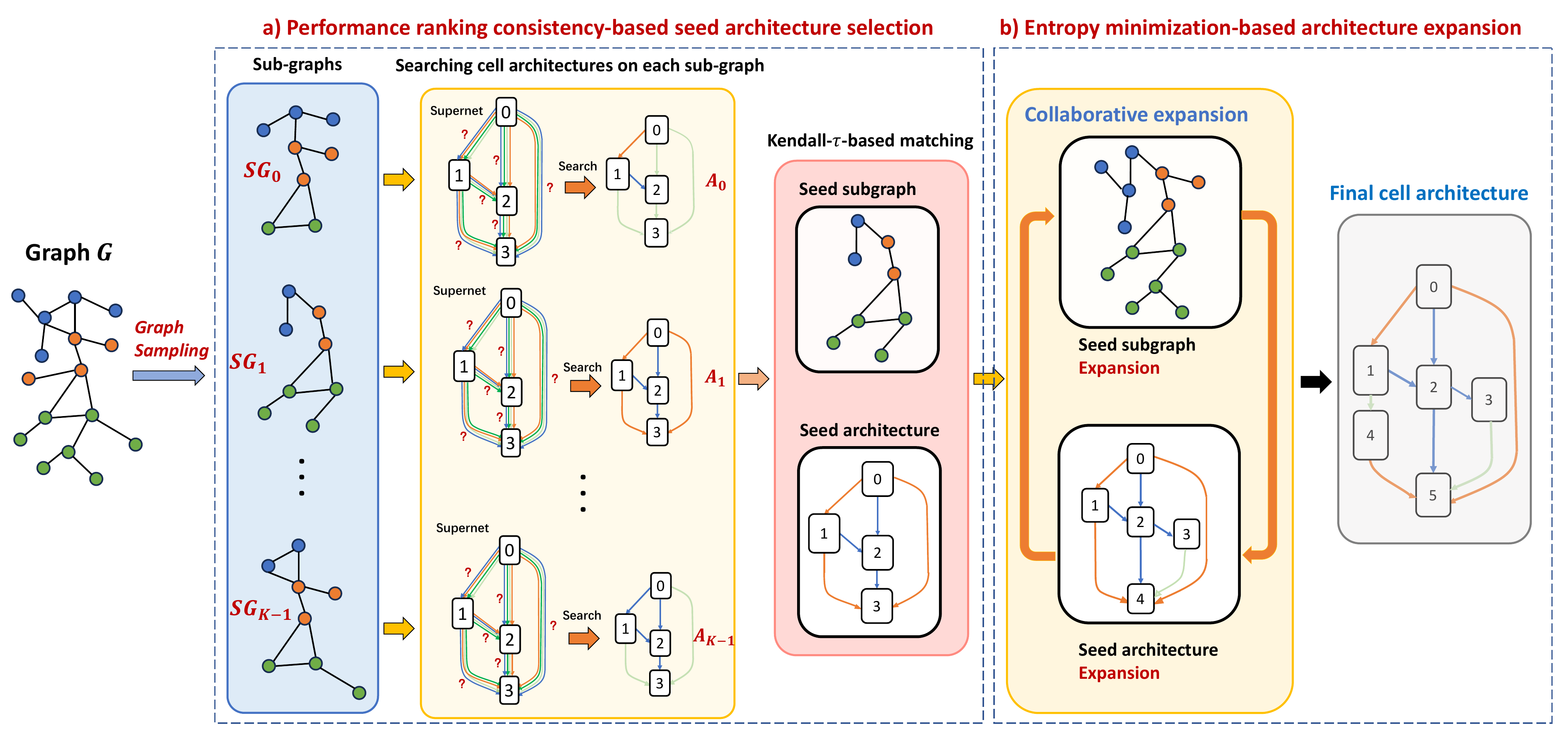}
\caption{The overall search framework of \SAGNAS{}, which consists of two stages: a) performance ranking consistency-based seed architecture selection; b) Entropy minimization-based seed architecture expansion.}
\label{fig:overview}
\end{figure*}

\subsection{Overall Search Framework}
Given the cell-based search space, we propose a seed architecture expansion-based search framework for large-scale GNAS.
Architecture expansion refers to the progressive splitting of nodes within the cell, resulting in an increasing number of nodes.
Unlike existing GNAS methods that directly search for the GNN architecture on large-scale graphs, we design a two-stage search strategy, which first identifies a highly promising seed architecture through subgraph sampling and matching, and then gradually expands the seed architecture to fit the original large-scale graph.  

Figure~\ref{fig:overview} illustrates the overall search framework of the proposed \SAGNAS{}, which consists of two stages, i.e., performance ranking consistency-based seed architecture selection and entropy minimization-based architecture expansion.

In the first stage, a set of subgraphs is sampled from the original large-scale graph using a light-weight and low-bias graph sampling method, e.g., node sampling in GraphSAINT~\cite{graphsaint}.
For each subgraph, we utilize an efficient gradient-based method to search the optimal cell architecture.
To determine the seed subgraph (i.e., the most consistent with the original graph) and the corresponding seed architecture, we propose a performance-driven matching strategy.
Specifically, each of the searched architectures is first evaluated on all subgraphs as well as the original graph.
We then construct performance sequences for all subgraphs and the original graph separately.
The length of performance sequence equals to the number of searched architectures, and each item in the sequence representing the validation performance of the searched architecture on the corresponding graph.
The weighted Kendall $\tau$ coefficient is employed to measure the ranking correlation between the performance sequences of each subgraph and the original graph.
The subgraph with the largest Kendall $\tau$ coefficient is selected as the seed subgraph, and the cell architecture searched on this subgraph is viewed as the seed architecture.

To enhance the expressive capabilities of the selected seed architecture on the original large-scale graph, we further iteratively expand the seed architecture and increase the architecture complexity.
In each iteration, we select the node with the highest entropy for splitting.
Subsequently, we employ a differentiable search method to identify the optimal local architecture for both the split node and the newly generated node.
The target of the localized architecture search is to find the optimal operations on the input and output edges of the two nodes.
Furthermore, during the search process, we collaboratively expand the scale of the seed subgraph to improve its adaptability to the original graph and to accommodate the increasing architecture complexity.



\subsection{Seed Architecture Selection}
\label{ss:seed_selection}
This stage first samples multiple subgraphs from the original large-scale graph and employs differentiable graph NAS for each subgraph.
Then, Kendall $\tau$ coefficient is used to measure the performance ranking consistency between each subgraph and the original graph. 
The searched architecture corresponding to the most consistent subgraph is selected as the seed architecture.

\subsubsection{Seed Architecture Generation with Differentiable GNAS}

To generate seed architecture for the large-scale graph, we employ a graph sampling strategy to sample multiple subgraphs and conduct GNN architecture search for each subgraph.
All searched architectures construct the candidate seed architecture pool, denoted as $\mathcal{A}$.
To improve the sampling efficiency while minimizing the node aggregation bias, we utilize the graph sampling algorithm GraphSAINT~\cite{graphsaint}.
Let $V$ and $V_s$ denote the node sets of the original graph $G$ and the sampled subgraph $SG$. $\widetilde{A}$ is the normalized adjacency matrix of $G$. 
The graph sampling algorithm GraphSAINT is not only light-weight, but also has the following good property.

\begin{prop}
~\cite{graphsaint}
$\zeta_v^{l+1}$ is an unbiased estimator of the aggregation of $v \in V_s$ in the $(l+1)^{th}$ GCN
layer, i.e., $\mathbb{E}{(\zeta_v^{l+1})} = \sum_{u \in V} \widetilde{A}_{u,v}(W^{l})^Th_{u}^{l}$, where $W^{l}$ and $h_{u}^{l}$ denote the weight and hidden embedding of the $l^{th}$ layer.
\end{prop}

Due to the low node aggregation bias, using GraphSAINT in \SAGNAS{} can achieve better performance than using the commonly-used cluster-based sampling algorithm~\cite{gcn-cluster} (Section~\ref{sss:sampling}).
Let ${SG}_i$ be the $i$-th subgraph, $A_i \in \mathcal{A} $ be the searched architecture on ${SG}_i$.
we leverage the differentiable architecture search strategy that relaxes the discrete choice of a specific operation to a softmax combination of all candidate operations.  
As introduced in Section~\ref{ss:space}, the GNN cell architecture is represented as a directed acyclic graph (DAG), containing two input nodes and $P$ intermediate nodes.
Considering the node pair of $N_i$ and $N_j$, denoted as $<i, j>$ for simplicity, there exists an edge $e_{i,j}$, directed from node $i$ to node $j$ ($i < j$). 
$e_{i,j}$ represents a GNN aggregation operation that transfers $h_{i}$ (i.e., the output of node $N_i$) to $\hat{h}_{i,j}$.

Let $\mathcal{O}$ denote the candidate aggregation operation set.
%
During the differentiable search process, the search space can be relaxed to be continuous and the operation in $e_{i,j}$ is viewed as a weighted mixture of candidate choices rather than an exclusive choice.
Each operation $o \in \mathcal{O}$, corresponds to a continuous weight coefficient $a^{(i,j)}_o$. 
Specifically, the output $\hat{h}_{i,j}$ can be calculated as follows:
\begin{equation}
    \hat{h}_{i,j} = {\textstyle \sum_{o\in \mathcal{O}}^{}\frac{exp(\alpha _{o}^{(i,j)} )}{ {\textstyle \sum_{o'\in \mathcal{O}}^{}}exp(\alpha _{o'}^{(i,j)}) } o(h_{i}) }.
\end{equation}

%
We divide ${SG}_i$ into the training set and the validation set.
Let $\mathcal{L}_{train}$ and $\mathcal{L}_{val}$ denote the training loss and validation loss, respectively. 
Since both losses are determined by the architecture parameters $\alpha$ and the weights $w$, the search objective is
a bi-level optimization problem. 
%
%
\begin{equation}
\begin{gathered}
\min_\alpha \mathcal{L}_{val}(w^*, \alpha) \\
s.t. \quad w^* = \text{argmin}_w \mathcal{L}_{train}(w, \alpha),
\end{gathered}
\label{differ-nas}
\end{equation}
where the upper-level optimization focuses on determining the optimal architecture parameters $\alpha$, while the lower-level optimization aims to identify the optimal weights $w$ in $A_i$.

It's worth noting that the architecture search process for $K$ subgraphs is inherently parallel.
Therefore, the search efficiency can be improved through  parallel implementation.

\subsubsection{Performance Ranking Consistency-based Subgraph Matching}
\label{sss:subgraph-matching}
We aim to select the most promising architecture from all the searched architectures as the seed architecture (i.e., $A_{seed}$).
To this end, we need to first identify the subgraph that most closely matches the original graph, which we refer to as the seed subgraph (i.e., $SG_{seed}$).
The architecture searched on $SG_{seed}$ is denoted as $A_{seed}$.

\begin{figure*}[htpb]
\centering
\includegraphics[width=0.85\textwidth]{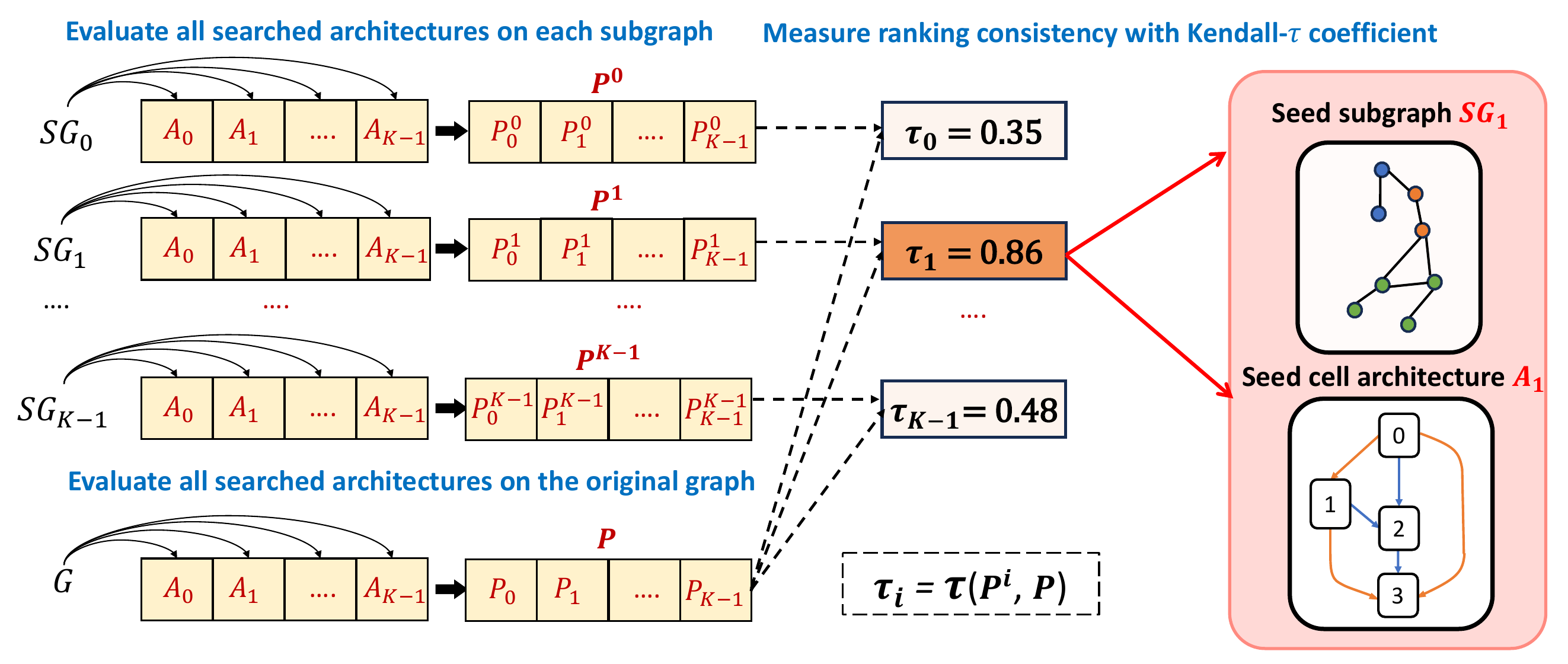}
\caption{Performance ranking consistency-based subgraph matching.}
\label{fig:match}
\end{figure*}

To measure the consistency between each sampled subgraph and the original graph, we propose a performance ranking consistency-based method.
As shown in Figure~\ref{fig:match}, the original graph $G$ and each subgraph maintains a performance sequence.
Performance evaluation is performed under the inductive node classification mode.
Let $P_k^i$ denote the prediction performance of architecture $A_{k}$ on the validation set of subgraph $SG_i$.
For subgraph $SG_i$, the corresponding performance sequence is $\textbf{P}^i = \{P_k^i \ | \ k \in [0,K-1]\}$.
Similarly, $P_k$ denotes the evaluation performance of architecture $A_{k}$ on the validation set of original graph $G$, and the performance sequence of graph $G$ is $\textbf{P} = \{P_k \ | \ k \in [0,K-1]\}$.
The more consistent the performance ranking between $\textbf{P}^i$ and $\textbf{P}$, the better $SG_i$ and $G$ match.
%

To quantify the similarity between the performance ordering of $\textbf{P}^i$ and $\textbf{P}$, we employ weighted Kendall $\tau$ ~\cite{Puka2011, SHIEH199817, 10027699}.
Specifically, Kendall $\tau$ is a statistical metric used to quantify the ranking consistency of two sequences. 
Let $\triangle^{SG_i}_{k,k'} = P_k^i - P_{k'}^i$ be the performance difference of $A_k$ and $A_k'$ on subgraph $SG_i$, and $\triangle^G_{k,k'} = P_k - P_{k'}$ be the performance difference of $A_k$ and $A_k'$ on the original graph $G$.
If $\triangle^{SG_i}_{k,k'} / \triangle^G_{k,k'}$ is negative, the rank-order is discordant and vice versa is concordant. 
The weighted Kendall $\tau$ coefficient between $\textbf{P}^i$ and $\textbf{P}$ is defined as follows:

\begin{align}
\tau_i &= \tau(\textbf{P}^i, \textbf{P}) = \frac{\sum_{k=0}^{K-2}\sum_{k'=k+1}^{K}w^{SG_i,G}_{k,k'}}{\sum_{k=0}^{K-2}\sum_{k'=k+1}^{K}\|w^{SG_i,G}_{k,k'}\|}, \ \ \text{where} \notag \vspace{1ex}  \\
& w^{SG_i,G}_{k,k'}= \left\{ \begin{array}{lll}
    1 &   \text{if} &  \triangle^{SG_i}_{k,k'} = \triangle^G_{k,k'} = 0 \vspace{1ex}  \\
    \triangle^{SG_i}_{k,k'} / \triangle^G_{k,k'}  & \text{else  if}   &   \| \triangle^{SG_i}_{k,k'} \| \leq \| \triangle^G_{k,k'} \| \vspace{1ex} \\
     \triangle^G_{k,k'} / \triangle^{SG_i}_{k,k'} & \text{else; i.e.,}  & \| \triangle^{SG_i}_{k,k'} \| > \| \triangle^G_{k,k'} \|.
\end{array}
\right.
\end{align}

%
%

Kendall $\tau$ quantifies the concordance/discordance between pairwise architecture performances. 
When the rank-orders are discordant, or when they are concordant but the $\triangle$-differences are
disproportionate (i.e., $w^{SG_i,G}_{k,k'}$ is near-zero), Kendall $\tau$ becomes smaller. 
If the rand-orders of $\textbf{P}^i$ and $\textbf{P}$ are mostly consistent, $\tau_i$ is close to 1. Conversely, if they are mostly opposite, $\tau_i$ tends to be -1.
Thus, higher $\tau_i$ indicates that performance ranking is more consistent and $SG_i$ and $G$ are more similar.

As a result, we can select the subgraph with the highest Kendall $\tau$ as the seed subgraph, and the corresponding searched architecture as the seed architecture.
The detailed process for searching the seed architecture is outlined in Algorithm~\ref{alg:selection}.

\begin{algorithm} 
	\caption{Seed Architecture Selection} 
	\label{kendall} 
 \begin{flushleft}
        \textbf{INPUT:} The graph $G$, the graph sampler $\pi(G)$\\
        \textbf{OUTPUT:} The seed subgraph $SG_{seed}$ and seed architecture $A_{seed}$
\end{flushleft}
\begin{algorithmic}[1]
        \STATE  sample $K$ times on $G$ with $\pi(G)$ and get the subgraph set $SG$
        \STATE initialize the candidate seed architecture pool $\mathcal{A}$
        \FOR   {$SG_{i}$ in $SG$}
        \STATE construct the cell-based supernet as the search space
        \STATE run differential graph NAS on $SG_{i}$ according to Equation~\ref{differ-nas} 
        \STATE add $A_{i}$ searched on $SG_{i}$ into $\mathcal{A}$ 
        \ENDFOR
        \STATE evaluate all candidate architectures in $\mathcal{A}$ on the original graph and get the performance sequence $\textbf{P}$
        \FOR {$SG_{i}$ in $SG$}
           \STATE evaluate all candidate architectures in $\mathcal{A}$ on $SG_{i}$ and get the performance sequence $\textbf{P}^i$
           \STATE calculate the Kendall $\tau$ coefficient between $\textbf{P}^i$ and $\textbf{P}$
        \ENDFOR 
        \STATE select the subgraph with the largest Kendall $\tau$ as the seed subgraph $SG_{seed}$ 
        \STATE select the architecture searched on $SG_{seed}$ as the seed architecture $A_{seed}$
        \STATE return $SG_{seed}$ and $A_{seed}$
	\end{algorithmic} 
 \label{alg:selection}
\end{algorithm}

\subsection{Seed Architecture Expansion}

\label{section:Entropy}
Although the seed architecture $A_{seed}$ searched on the seed subgraph $SG_{seed}$ shows promising performance, it still suffers from two limitations.
First, from an architectural perspective, the parameter scale of $A_{seed}$ is relatively small, leading to the underfitting problem when applying to original large-scale graph.
Second, from a data perspective, the differences in feature distribution and topological characteristics between the subgraph and the entire graph result in consistency collapse issues when the seed architecture is directly transferred to the original graph~\cite{hu2021open}.
Thus, to adapt to the original large graph, we propose an architecture and subgraph collaborative expansion method to improve the overall performance.

Specifically, we propose an entropy minimization based expansion strategy that first selects an intermediate node with the highest entropy from the cell architecture and then performs node splitting.
To further search for the optimal local architecture for both the split node and the newly generated node, we employ a differentiable search method that incorporates subgraph expansion and entropy minimization regularization.

\subsubsection{Entropy Minimization based Architecture Expansion}

We represent the probability distribution of the operations in edge $e_{i,j}$ as $\textbf{P}^{(i,j)}$, where the probability of selecting operation $o$ is defined as:

\begin{equation}
p_o^{(i,j)} = \frac{exp(\alpha_o^{(i,j)})}{\sum_{o^{\prime}\in \mathcal{O}}exp(\alpha_{o'}^{(i,j)})}.
\end{equation}
Thus, the output $\hat{h}_{i,j}$ is equivalent to the expected value of the outputs generated by all available operations.
\begin{equation}
    \hat{h}_{i,j} = \mathbb{E}_{p_o \in \textbf{P}^{(i,j)}}[p_o \cdot o(h_{i})].
\end{equation}

After the differentiable search process, the continuous choices need to be discretized. 
The operation with the largest weight coefficient, i.e., the highest selection probability in $e_{i,j}$ will be retained. 
%
%
Ideally, the optimized distribution $\textbf{P}^{(i,j)}$ should converge to an extreme long-tailed distribution, which implies that the probability of a specific operation approaches 1, the other tends to 0. 
Such convergence minimizes the loss of information resulting from the discarding of aggregation operations.
To measure the certainty and reliability of operations in edge $e_{i,j}$, we introduce the definition of edge entropy.

\begin{myDef}
Edge entropy $ETP^{(i,j)}_{e}$ is used to quantify the uncertainty of the probability distribution of the operations in $e_{i,j}$. The edge entropy is formally defined as:
\begin{equation}
    ETP^{(i,j)}_{e}={-\sum_{o\in \mathcal{O}}p_o^{(i,j)}\ln\left(p_o^{(i,j)}\right)}.
\end{equation}
\end{myDef}
A higher edge entropy indicates a more uniform probability distribution among operations and greater uncertainty, suggesting that more information is lost when discarding operations during the discretization process.
Since the fundamental building block of the cell architecture is the node, the architecture expansion should be performed at the node level.
To identify nodes with low certainty as well as poor stability for splitting, we further introduce node entropy based on the edge entropy.

\begin{myDef}
Node entropy $ETP^{j}_{n}$ is used to quantify the stability of node $j$ in the cell architecture. 
Let $\mathcal{I}_j$ denote the input edge set of node $j$.
The node entropy is the average of the entropies of all input edges, which is formally defined as:
\begin{equation}
    ETP^{j}_{n}=\frac{\sum_{e_{i,j}\in \mathcal{I}_j}ETP^{(i,j)}_{e}}{|\mathcal{I}_j|}.
    \label{equ:node_etp}
\end{equation}
\end{myDef}

%
For each intermediate node, node entropy serves as a measure of the uncertainty and stability associated with the messages it receives.
In each iteration of architecture expansion process, for node $j$ with the highest node entropy, we perform node splitting on node $j$ and introduce a new node to share the information load with node $j$. 
On one hand, we can reduce the entropy of node $j$ and improve the overall stability of the cell architecture.
On the other hand, the architecture complexity becomes higher through node splitting, and its representation capability increases.

\begin{figure*}[t]
\centering
\includegraphics[width=1\linewidth]{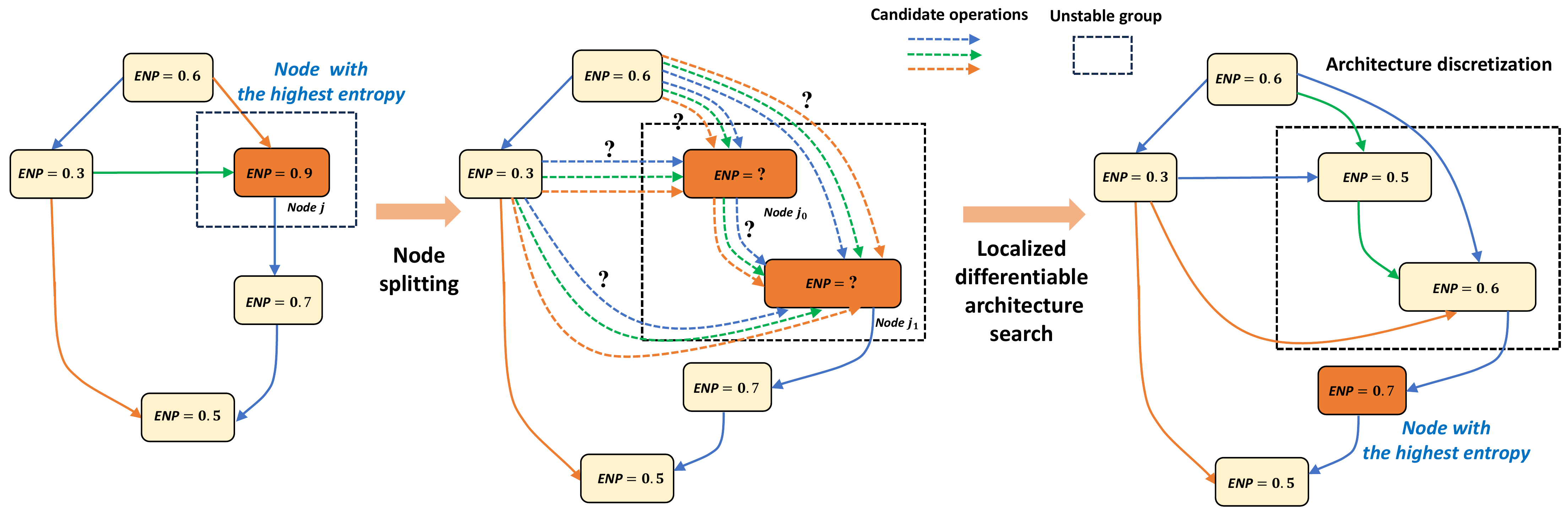}
\caption{The architecture expansion based on entropy minimization and localized differentiable architecture search.}
\label{fig:entropy}
\vspace{-1ex}
\end{figure*}

%
Specifically, Figure~\ref{fig:entropy} illustrates the architecture expansion process through node splitting.
Assume that node $j$ has the highest node entropy and thus becomes the target to be split.
Let node $j_0$ and node $j_1$ be the two nodes split from node $j$.
Both node $j_0$ and node $j_1$ inherit the input edge set of node $j$.
For node $j_1$, there exists an input edge from node $j_0$.
Thus, the input edge sets of node $j_0$ and node $j_1$ are as follows:
\begin{equation}
    \mathcal{I}_{j_0} = \mathcal{I}_{j}, \quad \mathcal{I}_{j_1} = \mathcal{I}_{j} \cup \{e_{{j_0},{j_1}}\}.
\end{equation}

For the input edge of node $j_0$ and node $j_1$, the candidate aggregation operation set is $\mathcal{O}$.
Therefore, the following objective is to search the optimal operation in each input edge.
To this end, we further propose a localized differential architecture search strategy with entropy minimization regularization.

\subsubsection{Localized Differential Architecture Search with Entropy Minimization Regularization}
As mentioned in Section~\ref{ss:seed_selection}, the seed architecture $A_{seed}$ is searched on the seed subgraph $SG_{seed}$.
As the architecture continues to expand, the GNN model becomes increasingly complex, which may lead to overfitting on $SG_{seed}$.
To mitigate this issue, we propose a collaborative expansion method that combines subgraph and architecture expansion.
By expanding subgraphs, we can more accurately capture the characteristics of the original graph, leading to improved performance of the searched architecture.
Specifically, for each node in the subgraph, we randomly add $M$ 1-hop neighbors from the original graph.


Furthermore, to reduce the overall node entropy and improve the stability of the searched architecture, we introduce an additional regularization term designed to constrain the overall entropy of the architecture.
The overall entropy is defined as the average of all node entropies.
Therefore, the regularization term is as follows:
 \begin{equation}
     \mathcal{L}_{reg} = \frac{1}{P}\sum \limits _{j\in P} ETP_n^j \ ,
 \end{equation}
where $P$ is total number of intermediate nodes in the cell architecture. 
The search process can be formulated as a bi-level optimization problem, with $\lambda$ representing a hyperparameter to adjust the weight of $\mathcal{L}_{reg}$.

\begin{equation}
\begin{gathered}
\min_{\hat{\alpha}} \mathcal{L}_{val}(w^*, \hat{\alpha}, SG_{seed}^i) + \lambda \mathcal{L}_{reg}\\
s.t. \quad w^* = \text{argmin}_w \mathcal{L}_{train}(w, \hat{\alpha}, SG_{seed}^i).
\end{gathered}
\end{equation}

To elaborate on the overall workflow of subgraph and architecture collaborative expansion, we provide additional pseudo-code in Algorithm~\ref{workflow}.

\begin{algorithm}
\caption{Seed Architecture Expansion} 
	\label{workflow} 
 \begin{flushleft}
        \textbf{INPUT:} The seed architecture $A_{seed}$, the seed subgraph $SG_{seed}$, the graph $G$, the graph sampler $\pi(G)$, loss function $\mathcal{L}$, expansion stopping condition $\Phi$\\
        \textbf{OUTPUT:} The final searched architecture $A_{final}$
\end{flushleft}
    \begin{algorithmic}[1]
        \STATE run Algorithm~\ref{alg:selection} to obtain the seed subgraph $SG_{seed}$ and the seed architecture $A_{seed}$.
       
        \WHILE {not $\Phi$}  

        \STATE calculate node entropy for each intermediate node according to Equation~\ref{equ:node_etp}
        
        \STATE select the node with the highest node entropy 

        \STATE perform node splitting and expand the seed  architecture $A_{seed}$

        \STATE expand the seed graph $SG_{seed}$ from $G$ with $\pi(G)$

        \STATE split the training and validation sets from $SG_{seed}$

        \STATE fix the localized architecture parameters $\hat{\alpha}$, calculate $\nabla_w \mathcal{L}_{train}(w, \hat{\alpha})$, and update weights $w$

        \STATE fix $w$, calculate $\nabla_{\hat{\alpha}} (\mathcal{L}_{val}(w, \hat{\alpha}) + \lambda \mathcal{L}_{reg} )$, and update architecture parameters $\hat{\alpha}$
        
        \ENDWHILE 

        \STATE return the final architecture $A_{final}$
	\end{algorithmic} 
\end{algorithm}

\subsubsection{Analysis on Node Splitting}
\label{sss:analysis_split}
For the node with the highest entropy, splitting it and introducing new nodes to carry its original high-density information flow help to reduce its entropy. 
As shown in Figure~\ref{fig:entropy}, we collectively refer to the split node and the newly added nodes as the \emph{unstable group} (UG for short). Before splitting, the unstable group only contains node $j$. 
After splitting, the unstable group contains node $j_0$ and node $j_1$.
Let \( ETP_{UG} \) represent the sum of the entropies of the nodes in the unstable group.
Intuitively, the relationship before and after the splitting is given by: \( ETP_{UG}^{before} \ge ETP_{UG}^{after} \). 
Based on this intuitive conjecture, we propose the following proposition:

\begin{prop}
\label{proposition2}
Let \( N \) be the number of cell nodes in the initial state of the $i$-th split of seed architecture \( A^i_{\text{seed}} \), and let \( ETP_{\text{stable}} \) represent the sum of the entropies of all nodes in \( A^i_{\text{seed}} \) except for those in the unstable group. For \( A^i_{\text{seed}} \), the architecture entropy after node splitting and localized differentiable search decreases, which can be formally expressed as:
 \begin{equation}
\frac{ETP_{\text{stable}} + ETP_{UG}^{before}}{N} \ge \frac{ETP_{\text{stable}} + ETP_{UG}^{after}}{N+1}.
 \end{equation}
\end{prop}
Experimental results in Section~\ref{sss:effectivenss_expansion} further validates the above conjecture and proposition, indicating the effectiveness of choosing the node with the highest
entropy for splitting.

\section{Experiments}
In this section, we conduct extensive experiments on large-scale graph datasets to answer the following questions:

\begin{itemize}
    \item \textbf{RQ1}: How do the architectures discovered by \SAGNAS{} compared to handcrafted architectures and the ones discovered by other GNAS methods? 
    \item \textbf{RQ2}: How does the search efficiency of \SAGNAS{} compared to other GNAS methods?
    \item \textbf{RQ3}: What are the impacts of different design strategies in \SAGNAS{} influence performance?
     \item \textbf{RQ4}: How does the hyperparamters in \SAGNAS{} influence its performance?
    \item \textbf{RQ5}: 
    What insights can we draw from the searched architectures?
\end{itemize}

\subsection{Experiment Setup}
\subsubsection{Datasets}
To verify the effectiveness and scalability of the proposed \SAGNAS{}, we conduct experiments on five graph datasets at different scales. 
The node sizes range from $10^4$ to $10^8$.
Table~\ref{tab:dataset} presents the statistics of the five graph datasets. 
The graph learning task focuses on node classification.
Two of these datasets are from the GNN benchmark: Coauthor CS and Coauthor Physics, while the other three are from the Open Graph Benchmark (OGB)~\cite{hu2021open}: ogbn-arxiv, ogbn-products, and ogbn-papers 100M.

\begin{itemize}
\item Coauthor CS and Coauthor Physics are co-authorship graphs in the fields of computer science and physics, where nodes represent individual authors, edges indicate co-authorship on research papers, and class labels indicate the most active research fields for each author.
\item OGB provides a comprehensive collection of realistic, multi-scale, and multi-task graph datasets.
We choose three datasets of varying scales under the Node Property Prediction classification of the OGB benchmark.
Ogbn-arxiv is relatively small in the OGB benchmark, representing a directed graph of citations between all Computer Science (CS) arXiv papers indexed by MAG~(Microsoft Academic Graph)~\cite{Wang2020MicrosoftAG}.
In contrast, ogbn-products and ogbn-papers 100M contain more than millions of nodes.
ogbn-products is based on the Amazon product co-purchasing network, which is an undirected and unweighted graph.
ogbn-papers 100M is the largest, comprising over 111 million nodes and 1 billion edges.
%
\end{itemize}

\begin{table}[t]
\caption{Dataset Statistics}
\scalebox{1}{
\resizebox{\linewidth}{!}{%
\begin{tabular}{@{}lcccc@{}}
\toprule
\textbf{Dataset} & \textbf{Nodes} & \textbf{Edges} & \textbf{Features} & \textbf{Classes} \\ \midrule
CS               & 18,333         & 81,894         & 6,805             & 15               \\ \midrule
Physics          & 34,493         & 247,962        & 8,415             & 5                \\ \midrule
ogbn-arxiv       & 169,343        & 1,166,243      & 128               & 40               \\ \midrule
ogbn-products    & 2,449,029      & 61,859,140     & 100               & 47               \\ \midrule
ogbn-papers 100M  & 111,059,956    & 1,615,685,872  & 128               & 172              \\ \bottomrule
\end{tabular}
}
}
\label{tab:dataset}
\end{table}

\subsubsection{Baselines}
We compare \SAGNAS{} with representative human-designed state-of-the-art GNN architectures and seven GNAS baselines.
\begin{itemize}

\item The human-designed GNN architectures include GCN~\cite{kipf2017gcn}, GAT~\cite{velivckovic2018gat}, GraphSAGE~\cite{hamilton2017sage}, and GIN~\cite{pmlr-v162-wang22am}.
\item The GNN NAS baselines include GraphNAS~\cite{gao2020graphnas}, AutoGNAS~\cite{AutoGNAS}, AutoGraph~\cite{AutoGraph}, SGAS~\cite{sgas}, EGAN~\cite{EGAN}, PSP~\cite{psp}, and GAUSS~\cite{gauss}. 
\end{itemize}
For each human-designed model and GNAS method, we rerun each approach five times and report the mean and variance of the performance to account for random factors.
Since GAUSS is not open-source, we directly adopt the performance presented in the original paper~\cite{gauss}.

\subsubsection{Training Details}
%
We implement all models with PyG and train them using the Adam optimizer.
To ensure low bias compared to the original graph, we employ GraphSAINT~\cite{graphsaint} for subgraph sampling.
We also evaluate and compare different sampling strategies in Section~\ref{In-Depth}.
We initially set the number of intermediate nodes in the cell to 3. 
The number of intermediate nodes gradually increases during architecture expansion. 
%
We verify the influence of the number of intermediate nodes on different graph datasets in Section ~\ref{nodestudy}. 
Additionally, the details of hyperparameters such as the number of stacked cells, the number of intermediate nodes in the cell, and the number of sampled subgraphs, and the hidden embedding size are illustrated in Table~\ref{parameter}. 
All experiments are conducted on a NVIDIA V100 GPU with 32 GB memory.

\begin{table*}[t]
\caption{Detailed hyperparameters settings.}
\centering
\label{parameter}
\resizebox{0.8\linewidth}{!}{%
\begin{tabular}{@{}lcccccc@{}}
\toprule
\textbf{Dataset} & \textbf{\begin{tabular}[c]{@{}c@{}} \# stacked \\ cells \end{tabular}} & \textbf{\begin{tabular}[c]{@{}c@{}} \# intermediate nodes \\ in the cell \end{tabular}} & \textbf{\begin{tabular}[c]{@{}c@{}} \# sampled \\ subgraphs\end{tabular}} & \textbf{\begin{tabular}[c]{@{}c@{}}hidden \\ embedding size\end{tabular}} & \textbf{\begin{tabular}[c]{@{}c@{}}learning \\ rate\end{tabular}} & \textbf{\begin{tabular}[c]{@{}c@{}}weight \\ decay\end{tabular}} \\ \midrule
CS               & 5                                                                  & 6                                                                   & 9                                                                      & 64                    & 2e-3                   & 3e-4                  \\
Physics          & 5                                                                  & 6                                                                   & 10                                                                     & 64                    & 2e-3                   & 3e-4                  \\
Arxiv            & 6                                                                  & 7                                                                   & 8                                                                      & 128                   & 5e-4                   & 3e-4                  \\
Products         & 6                                                                  & 8                                                                   & 9                                                                      & 128                   & 5e-4                   & 3e-4                  \\
Papers100M       & 6                                                                  & 8                                                                   & 9                                                                      & 128                   & 5e-4                   & 3e-4                  \\ \bottomrule
\end{tabular}
}
\end{table*}
%
%

\subsection{Performance Comparison (RQ1)}
Table~\ref{result} shows the performance comparison between \SAGNAS{} and other baselines including SOTA human-designed GNN architectures and existing GNAS methods. 
The performance comparison is conducted on five datasets at different scales.
In line with GAUSS~\cite{gauss}, we exclude all training tricks to ensure a focused evaluation of the core performance of each method.
As shown in Table~\ref{result}, \SAGNAS{} consistently outperforms all human-designed models. 
Additionally, when compared to other GNAS methods, \SAGNAS{} demonstrates superior performance across all datasets, highlighting the effectiveness and scalability of the proposed approach.

Note that when applying GNAS methods on large-scale graphs (i.e., when the number of nodes exceeds $10^6$), such as ogbn-products and ogbn-papers 100M, most GNAS methods become less competitive due to computational and memory issues (out-of-time or out-of-memory). 
GraphNAS, based on reinforcement learning, requires over 1 GPU day on the ogbn-products and ogbn-papers 100M datasets.
The huge time overhead limits the application of GraphNAS on large-scale graphs.
%
Although differentiable search methods (e.g., AutoGNAS, SGAS, and EGAN) and the progressive pruning-based method PSP improve search efficiency to some extent, they still struggle with large-scale graphs due to GPU memory constraints. 
In comparison to GAUSS, the SOTA large-scale GNAS method, \SAGNAS{} not only achieves better performance on ogbn-products and ogbn-papers 100M but also demonstrates superior search efficiency (Section~\ref{ss-searchefficiency}).

\begin{table*}[ht]
\centering
\caption{Performance comparison with the SOTA human-designed GNN architectures and the existing GNAS methods. We report both the validation and the test accuracy over 5 runs with different seeds. OOT means out-of-time (cannot converge within 1 single GPU day), while OOM means out-of-memory (cannot run on a Tesla V100 GPU with 32GB memory). The best results are in bold. * denotes the results reported in the original paper. }
\label{result}
\resizebox{\textwidth}{!}{%
\begin{tabular}{@{}lcccccccccc@{}}
\toprule
\multirow{2}{*}{\textbf{Methods}} &
  \multicolumn{2}{c}{\textbf{CS}} &
  \multicolumn{2}{c}{\textbf{Physics}} &
  \multicolumn{2}{c}{\textbf{Arxiv}} &
  \multicolumn{2}{c}{\textbf{Products}} &
  \multicolumn{2}{c}{\textbf{Papers100M}} \\ \cmidrule(l){2-11} 
 &
  valid &
  test &
  valid &
  test &
  valid &
  test &
  valid &
  test &
  valid &
  test \\ \midrule
GCN &
  $93.51_{0.30}$ &
  $92.97_{0.31}$ &
  $96.04_{0.18}$ &
  $95.94_{0.17}$ &
  $71.74_{0.29}$ &
  $69.98_{0.19}$ &
  $91.57_{0.04}$ &
  $75.83_{0.21}$ &
  $68.32_{0.11}$ &
  $65.06_{0.04}$ \\
GAT &
  $93.45_{0.21}$ &
  $93.02_{0.31}$ &
  $95.98_{0.17}$ &
  $95.86_{0.13}$ &
  $72.78_{0.09}$ &
  $71.65_{0.23}$ &
  $90.85_{0.07}$ &
  $78.46_{0.19}$ &
  $68.76_{0.12}$ &
  $65.66_{0.18}$ \\
GraphSAGE &
  $95.21_{0.11}$ &
  $94.77_{0.26}$ &
  $96.35_{0.20}$ &
  $96.07_{0.14}$ &
  $72.67_{0.18}$ &
  $71.46_{0.21}$ &
  $91.63_{0.07}$ &
  $78.19_{0.27}$ &
  $69.12_{0.14}$ &
  $66.97_{0.13}$ \\
GIN &
  $91.97_{0.21}$ &
  $92.07_{0.17}$ &
  $95.92_{0.18}$ &
  $95.89_{0.09}$ &
  $70.96_{0.21}$ &
  $69.97_{0.24}$ &
  $91.70_{0.11}$ &
  $78.41_{0.21}$ &
  $67.98_{0.14}$ &
  $64.98_{0.39}$ \\ \midrule
GraphNAS &
  $92.70_{0.12}$ &
  $92.54_{0.09}$ &
  $96.76_{0.10}$ &
  $93.87_{0.06}$ &
  $72.76_{0.15}$ &
  $71.70_{0.18}$ &
  OOT &
  OOT &
  OOT &
  OOT \\
AutoGNAS &
  $92.90_{0.10}$ &
  $92.78_{0.09}$ &
  $94.50_{0.21}$ &
  $94.16_{0.07}$ &
  $70.13_{0.07}$ &
  $70.09_{0.13}$ &
  OOM &
  OOM &
  OOM &
  OOM \\
AutoGraph &
  $92.59_{0.03}$ &
  $92.54_{0.09}$ &
  $94.22_{0.15}$ &
  $94.02_{0.10}$ &
  $72.12_{0.05}$ &
  $71.08_{0.21}$ &
  OOM &
  OOM &
  OOM &
  OOM \\
SGAS &
  $95.62_{0.06}$ &
  $95.44_{0.06}$ &
  $96.44_{0.10}$ &
  $96.50_{0.07}$ &
  $72.38_{0.11}$ &
  $71.34_{0.25}$ &
  OOM &
  OOM &
  OOM &
  OOM \\
EGAN &
  $95.60_{0.10}$ &
  $95.43_{0.05}$ &
  $96.39_{0.18}$ &
  $96.45_{0.19}$ &
  $72.91_{0.25}$ &
  $71.75_{0.35}$ &
  OOM &
  OOM &
  OOM &
  OOM \\
PSP &
  $95.34_{0.08}$ &
  $95.28_{0.07}$ &
  $96.15_{0.14}$ &
  $96.67_{0.17}$ &
  $72.84_{0.23}$ &
  \multicolumn{1}{l}{$72.22_{0.14}$} &
  OOM &
  OOM &
  OOM &
  OOM \\
GAUSS* &
  $96.08_{0.11}$ &
  $96.49_{0.14}$ &
  $96.79_{0.06}$ &
  $96.76_{0.08}$ &
  $73.63_{0.10}$ &
  \multicolumn{1}{l}{$72.35_{0.21}$} &
  $91.60_{0.12}$ &
  $81.26_{0.36}$ &
  $70.57_{0.07}$ &
  $67.32_{0.18}$ \\
\textbf{\SAGNAS{}} &
  \textbf{96.78$_{0.08}$} &
  \textbf{96.72$_{0.11}$} &
  \textbf{96.83$_{0.04}$} &
  \textbf{96.89$_{0.10}$} &
  \textbf{74.24$_{0.11}$} &
  \textbf{73.77$_{0.10}$} &
  \textbf{91.81$_{0.21}$} &
  \textbf{81.64$_{0.21}$} &
  \textbf{70.77$_{0.13}$} &
  \textbf{67.79$_{0.11}$} \\ \bottomrule
\end{tabular}%
}
\end{table*}

\begin{table}[t]
    \caption{Search time (GPU hours) of each GNAS method on a Tesla V100 GPU with 32GB memory.}
    \resizebox{\linewidth}{!}{
    \begin{tabular}{lccccc}
        \toprule
        \textbf{Methods} & \textbf{CS} & \textbf{Physics} & \textbf{Arxiv} & \textbf{Products} & \textbf{Papers100M} \\ 
        \midrule
        GraphNAS        & 7.5          & 9.2           & 13.0          & OOT              & OOT                \\ 
        AutoGraph       & 1.41         & 3.25          & 7.8           & OOM              & OOM                \\ 
        AutoGNAS        & 1.27         & 2.43          & 6.97          & OOM              & OOM                \\
        PSP             & 1.25         & 3.68          & 8.23          & OOM              & OOM                \\ 
        SGAS            & 0.34         & 0.66          & 3.0           & OOM              & OOM                \\ 
        GAUSS           & /            & /             & /             & /                & $\approx$ 24       \\ 
        \midrule
        \textbf{\SAGNAS{}} & \textbf{0.043}  & \textbf{0.13}  & \textbf{0.33}  & \textbf{0.67}   & \textbf{8.46}      \\ 
        \textbf{Speedup}   & \textbf{7.9$\times$} & \textbf{5.1$\times$} & \textbf{9.1$\times$} & \textbf{/} & \textbf{2.8$\times$} \\ 
        \bottomrule
    \end{tabular}
    }
    \label{timecost} 
\end{table}

\subsection{Search Efficiency Comparison (RQ2)}
\label{ss-searchefficiency}
Besides evaluating prediction performance, we further evaluate the search efficiency of each GNAS method.

\begin{table}[htbp]
\caption{Performance and search efficiency of parallel \SAGNAS{} on ogbn-products and ogbn-papers100M.}
\resizebox{\linewidth}{!}{
\begin{tabular}{@{}lcccc@{}}
\toprule
\multirow{2}{*}{\textbf{Device}} & \multicolumn{2}{c}{\textbf{Products}} & \multicolumn{2}{c}{\textbf{Papers100M}} \\ \cmidrule(l){2-5} 
 &
  \begin{tabular}[c]{@{}c@{}}Search time\\ (GPU hours)\end{tabular} &
  \begin{tabular}[c]{@{}c@{}}Test Accuracy\\ (\%)\end{tabular} &
  \begin{tabular}[c]{@{}c@{}}Search time\\ (GPU hours)\end{tabular} &
  \begin{tabular}[c]{@{}c@{}}Test Accuracy\\ (\%)\end{tabular} \\ \midrule
1      & 0.67                                               & $81.64_{0.21}$                                               & 8.46                                               & $67.79_{0.11}$                                               \\
2      & 0.33                                               & $81.24_{0.15}$                                               & 4.27                                               & $67.70_{0.01}$                                               \\
4      & 0.13                                               & $81.51_{0.13}$                                               & 2.31                                               & $67.72_{0.14}$                                               \\ \bottomrule
\end{tabular}
}
\label{parallel}

\end{table}

Since GAUSS~\cite{gauss} is not open-source, we refer to the results on Papers100M reported in the original paper.
%
%
Table~\ref{timecost} presents the search time of each GNAS method on a single GPU. 
\SAGNAS{} significantly reduces the search time, achieving better search efficiency.
For the graph datasets whose node sizes range from $10^4$ to
$10^6$, \SAGNAS{} reduces the search time from GPU hours to minutes, achieving 5.1$\times$-9.1$\times$ speedup.
For the largest graph, Papers100M, the search cost of GAUSS is approximately 1 GPU day, as the reinforcement learning-based search strategy in GAUSS requires a substantial number of architecture evaluations.
In contrast, the search time of \SAGNAS{} can be significantly reduced to 8.46 GPU hours, achieving a speedup of $2.8\times$.

Moreover, \SAGNAS{} can be naturally parallelized during the seed architecture selection stage (Section~\ref{ss:seed_selection}). 
We also implement parallel \SAGNAS{} and observe the search costs on two largest datasets. 
We conduct parallel \SAGNAS{} on 1, 2, and 4 GPUS, respectively. 
The results are presented in Table~\ref{parallel}.
It is evident that parallel \SAGNAS{} demonstrates good scalability as the number of GPUs increases.
In a 4-GPU configuration, the search time on Products and Papers100M can be reduced to 0.13 and 2.31 GPU hours, achieving speedups of $5.2\times$ and $3.7\times$, respectively.

\subsection{Ablation Studies (RQ3)}
\label{In-Depth}

\subsubsection{Effectiveness of performance ranking consistency-based seed architecture selection}
%
%
During the seed architecture selection stage, multiple subgraphs are sampled from the original large-scale graph.
The performance ranking consistency-based method is proposed to identify the subgraph that most closely aligns with the original large-scale graph.
The corresponding cell architecture searched on the selected subgraph is viewed as the seed architecture for further expansion. 
%
%
To measure performance ranking consistency, the Kendall $\tau$ coefficient is calculated between the performance sequences of each subgraph and the original graph.
The subgraph with the highest Kendall $\tau$ coefficient is selected as the seed subgraph (Section~\ref{sss:subgraph-matching}).

%
To verify the effectiveness of the performance ranking consistency-based seed architecture selection method, we compare \SAGNAS{} with three additional variants: (1) $M_{rand}$: randomly selecting an architecture from the candidate seed architecture pool $\mathcal{A}$ as the seed architecture; (2) $M_{highest}$: selecting the architecture with the highest validation accuracy on the original graph; (3) $M_{avr}$: selecting the architecture with the highest average validation accuracy on all subgraphs. 
%

\begin{table}[t]
\caption{Ablation study on the effectiveness of the performance ranking consistency-based seed architecture selection strategy. We compare three different seed architecture selection strategies and report the test accuracy on all datasets.}
\resizebox{\linewidth}{!}{%
\begin{tabular}{@{}llllll@{}}
\toprule
\textbf{Method}            & \textbf{CS}    & \textbf{Physics} & \textbf{Arxiv} & \textbf{Products} & \textbf{Papers100M}     \\ \midrule
$M_{rand}$ & $94.72_{0.39}$ & $96.12_{0.47}$ & $72.16_{0.41}$ & $81.24_{0.77}$ & $66.97_{0.36}$ \\
\begin{tabular}[c]{@{}l@{}}$M_{highest}$ \end{tabular} &
  $95.69_{0.17}$ &
  $96.17_{0.12}$ &
  $72.31_{0.11}$ &
  $81.12_{0.43}$ &
  $66.46_{0.24}$ \\
\begin{tabular}[c]{@{}l@{}}$M_{avr}$\end{tabular} &
  $95.13_{0.19}$ &
  $96.21_{0.09}$ &
  $72.15_{0.24}$ &
  $81.43_{0.26}$ &
  $67.06_{0.33}$ \\ \midrule
\textbf{Ranking consistency-based} &
  \textbf{96.72$_{0.11}$} &
  \textbf{96.89$_{0.09}$} &
  \textbf{73.77$_{0.10}$} &
  \textbf{81.64$_{0.21}$} &
  \textbf{67.79$_{0.11}$} \\ \bottomrule
\end{tabular}
}
\label{tab:kendallcom}
\end{table}

The comparison results with the other three strategies are listed in Table~\ref{tab:kendallcom}.
Selecting the seed architecture based on performance ranking consistency outperforms the other variants. 
In contrast, the random selection strategy performs poorly across all datasets, exhibiting significant performance variance that renders the subgraph matching process unstable.
The two additional strategies, $M_{highest}$ and $M_{avr}$, consider only the performance of candidate architectures on the original graph and on the subgraphs, respectively.
Both of them overlook the relationship between the subgraphs and the full graph, resulting in suboptimal model performance. 
%
By leveraging the performance ranking consistency-based subgraph matching strategy, the chosen subgraph (i.e., the seed subgraph) is the most relevant to the original large-scale graph.
Moreover, if the seed architecture continues to be expanded, more powerful representation learning capability will be obtained.

\begin{table}[t]
\caption{Ablation study on the effectiveness of the entropy minimization-based architecture expansion.
}
\resizebox{\linewidth}{!}{%
\begin{tabular}{@{}llllll@{}}
\toprule
\textbf{Method}            & \textbf{CS}    & \textbf{Physics} & \textbf{Arxiv} & \textbf{Products} & \textbf{Papers100M}     \\ \midrule
Without expansion                    & $93.51_{0.11}$ & $95.04_{0.20}$   & $69.81_{0.05}$ & $79.03_{0.27}$    & $65.15_{0.35}$ \\
Random expansion & $95.24_{0.09}$ & $96.41_{0.17}$   & $71.96_{0.18}$ & $81.07_{0.20}$    & $66.57_{0.12}$ \\ \midrule
\textbf{Entropy-based expansion}  & \textbf{96.72$_{0.11}$} &
  \textbf{96.89$_{0.09}$} &
  \textbf{73.77$_{0.10}$} &
  \textbf{81.64$_{0.21}$} &
  \textbf{67.79$_{0.11}$} \\ \bottomrule
\end{tabular}
}
\label{entropycom}
\end{table}


\subsubsection{Effectiveness of the entropy minimization-based architecture expansion}
\label{sss:effectivenss_expansion}
To evaluate the effectiveness of the entropy minimization-based architecture expansion, we compare \SAGNAS{} with two other variants: (1) Without augmentation: only involves the phase of selecting the seed architecture using the Kendall $\tau$-based subgraph matching method; (2) Random architecture expansion: randomly selects an intermediate node to augment the cell-based supernet. The results are shown in Table~\ref{entropycom}. 
We can observe that the proposed entropy minimization-based architecture expansion strategy demonstrates significant superiority over the two other expansion strategies across all datasets.
%

%
For the architecture expansion-free strategy, the seed architecture is determined following the Kendall $\tau$-based subgraph matching. 
The architecture is directly searched on the subgraph without any further architecture expansion.
While the seed architecture can best match the selected subgraph with superior performance, it cannot guarantee optimal performance when applied to the original large-scale graphs.
Additionally, compared to the strategy without expansion, the random expansion strategy can achieve better performance, but is still inferior to entropy-based expansion strategy.

\begin{figure}[t]
    \centering 
    \includegraphics[width=0.95\linewidth]{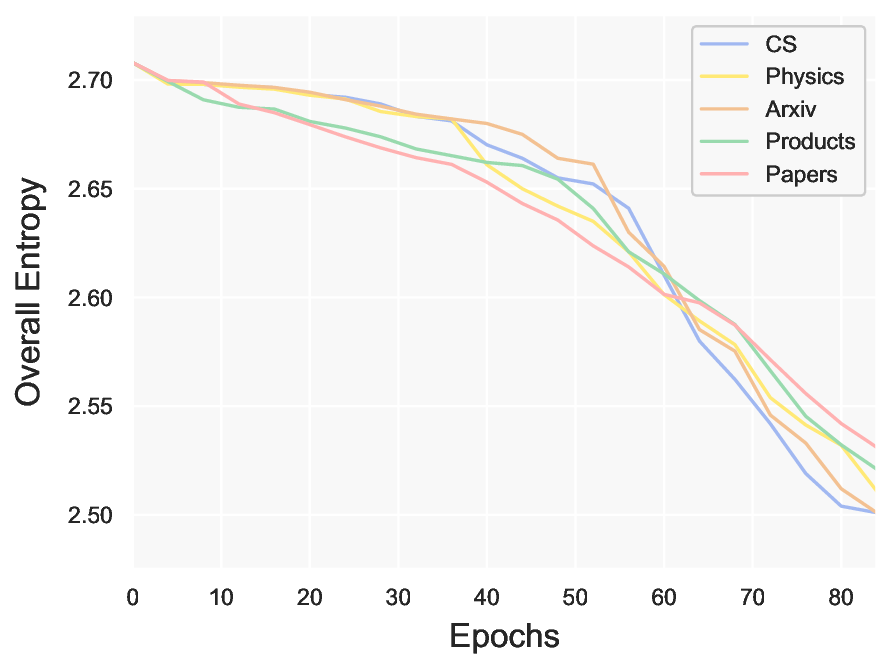}
    \caption{Trajectories of the overall entropy of
the cell architecture during the architecture expansion stage.}
    \label{fig:chartsofentropy}
\end{figure}

Moreover, we track the overall entropy of
the architecture during the expansion stage across five datasets in Figure~\ref{fig:chartsofentropy}.
As the architecture expansion progresses, the overall entropy continues to decrease, indicating that the cell-based architectures tend to be more stable following the expansion. 

Additionally, we track entropy change of the unstable group (defined in Section~\ref{sss:analysis_split}) during each split of the architecture expansion stage.
As shown in Figure~\ref{groupentropy}, we can observe that the architectural entropy of the unstable group decreases after node splitting and localized architecture search, which further validates Proposition 2, i.e., splitting the node with the
highest entropy and introducing new nodes is helpful to reduce the entropy and improve the stability.

\begin{table*}[htbp]
\centering
\caption{Trajectory of entropy change of the unstable group during each split of the architecture expansion stage.
Specifically, for $X \rightarrow Y$ in each split, $X$ and $Y$
represent the entropies of the unstable group before and after splitting, respectively. 
}
\resizebox{0.7\linewidth}{!}{%
\begin{tabular}{@{}lccccc@{}}
\toprule
\textbf{Split Index} &
  \multicolumn{1}{c}{CS} &
  \multicolumn{1}{c}{Physics} &
 Arxiv &
  \multicolumn{1}{c}{Products} &
  \multicolumn{1}{c}{Papers100M} \\ \midrule
1st Split & 2.68 $\rightarrow$ 2.49 $\downarrow$ & 2.65 $\rightarrow$ 2.50 $\downarrow$ & 2.65 $\rightarrow$  2.44  $\downarrow$ & 2.67 $\rightarrow$ 2.45 $\downarrow$ & 2.65 $\rightarrow$ 2.50 $\downarrow$ \\
2nd Split & 2.62 $\rightarrow$ 2.50 $\downarrow$ & 2.61 $\rightarrow$ 2.49 $\downarrow$ & 2.63 $\rightarrow$ 2.47 $\downarrow$ & 2.63 $\rightarrow$ 2.47 $\downarrow$ & 2.62 $\rightarrow$ 2.55 $\downarrow$ \\
3rd Split & 2.58 $\rightarrow$ 2.46  $\downarrow$ & 2.60 $\rightarrow$ 2.45 $\downarrow$ & 2.59 $\rightarrow$ 2.49 $\downarrow$ &  2.62 $\rightarrow$ 2.51 $\downarrow$ & 2.59 $\rightarrow$ 2.51 $\downarrow$ \\
4th Split & \multicolumn{1}{c}{-}   & \multicolumn{1}{c}{-}   & 2.60 $\rightarrow$ 2.55 $\downarrow$ & 2.58  $\rightarrow$ 2.46 $\downarrow$ & 2.59 $\rightarrow$ 2.47 $\downarrow$ \\
5th Split & \multicolumn{1}{c}{-}   & \multicolumn{1}{c}{-}   & -                       & 2.62 $\rightarrow$ 2.48  $\downarrow$ & 2.58 $\rightarrow$ 2.49 $\downarrow$ \\ \bottomrule
\end{tabular}%
}
\label{groupentropy}
\end{table*}

\begin{figure}[th]
    \centering  
    \includegraphics[width=0.95\linewidth]{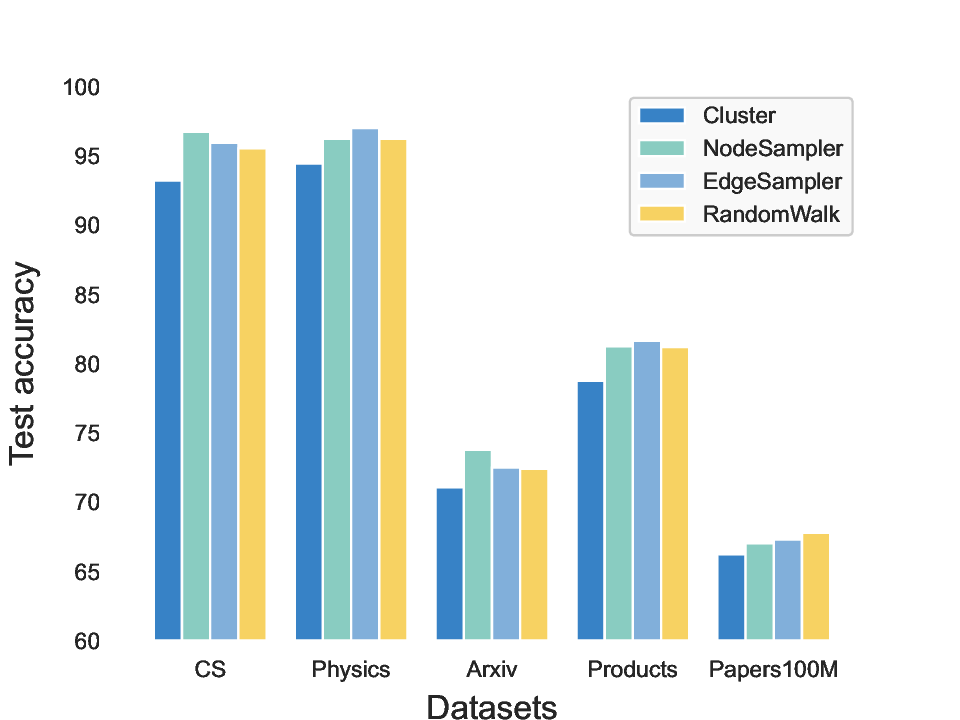}
    
    \caption{Performance comparison with different subgraph sampling strategies.}
    \label{samplingstrategy}
    %
\end{figure}

\subsubsection{Comparison between different subgraph sampling strategies}
\label{sss:sampling}
We also investigate the effects of different subgraph sampling strategies. 
Two widely adopted subgraph sampling techniques are evaluated, i.e., Cluster~\cite{gcn-cluster} and GraphSAINT~\cite{graphsaint}.
Cluster is a non-replacement sampling technique that selects clusters from the graph based on community detection.
GraphSAINT involves three distinct sampling strategies: node sampler, edge sampler, and random walk.
%
Figure~\ref{samplingstrategy} illustrates the performance comparison among four sampling strategies.

From the results in Figure~\ref{samplingstrategy}, it can be observed that the three sampling strategies of GraphSAINT demonstrate distinct strengths and weaknesses across different datasets, with performance discrepancies remaining within 1\%.  
Moreover, all three strategies outperform the Cluster sampling strategy.
This is because Cluster only captures the topological structure in the entire graph by sampling several clusters, leading to a significant deviation from the complete dataset.
As a result, Cluster provides weaker correlations with the complete graph, rendering it less effective for transferring architectures from subgraphs to the original large-scale graph.
\begin{table*}[htbp]
\caption{Compatibility with other training techniques. We report the test accuracy [\%] on five datasets. The underline indicates the performance of the best method that integrates all training techniques.}
\resizebox{\textwidth}{!}{%
\begin{tabular}{@{}lccccccccccccccc@{}}
\toprule
\multirow{2}{*}{\textbf{Method}} &
  \multicolumn{3}{c}{\textbf{CS}} &
  \multicolumn{3}{c}{\textbf{Physics}} &
  \multicolumn{3}{c}{\textbf{Arxiv}} &
  \multicolumn{3}{c}{\textbf{Products}} &
  \multicolumn{3}{c}{\textbf{Papers100M}} \\ \cmidrule(l){2-16} 
                                     & GCN   & GAT   & SA-NAS & GCN   & GAT   & SA-NAS & GCN   & GAT   & SA-NAS & GCN   & GAT   & SA-NAS & GCN   & GAT   & SA-NAS \\ \midrule
Plain                                & 92.97 & 93.02 & \textbf{96.72}  & 95.94 & 95.86 & \textbf{96.89}  & 69.98 & 71.65 & \textbf{74.24}  & 75.83 & 78.46 & \textbf{81.64}  & 65.06 & 65.66 & \textbf{67.79}  \\
Plain + Node2Vec                     & 93.54 & 93.84 & \textbf{96.88}  & 96.22 & 96.11 & \textbf{97.13}  & 71.26 & 72.05 & \textbf{74.32}  & 76.57 & 79.25 & \textbf{81.94}  & 65.55 & 66.32 & \textbf{67.99}  \\
Plain + Node2Vec + label             & 93.99 & 94.41 & \textbf{97.02}  & 96.64 & 96.58 & \textbf{97.30}  & 72.03 & 72.55 & \textbf{74.35}  & 76.95 & 80.38 & \textbf{82.44}  & 66.21 & 66.89 & \textbf{68.13}  \\
Plain + Node2Vec + label + C\&S      & 94.21 & 94.98 & \textbf{97.22}  & 96.87 & 96.81 & \textbf{97.37}  & 72.97 & 73.01 & \textbf{74.59}  & 77.75 & 81.28 & \textbf{83.01}  & 66.81 & 67.34 & \textbf{68.24}  \\
Plain + Node2Vec + label + C\&S + KD & 94.89 & 95.23 & \underline{\textbf{97.35}}  & 97.03 & 97.02 & \underline{\textbf{97.49}}  & 73.13 & 73.42 & \underline{\textbf{74.63}}  & 78.17 & 81.92 & \underline{\textbf{83.64}}  & 67.39 & 67.45 & \underline{\textbf{68.33}}  \\ \bottomrule
\end{tabular}%
}
\label{tbl:trick}
\end{table*}

\subsubsection{Compatibility with other techniques}
We demonstrate that the performance of searched architectures can be further improved by integrating additional training techniques. 
We employ four widely adopted techniques specifically designed to enhance node classification performance: Node2vec~\cite{node2vector}, Label Reuse~\cite{wang2021bagtricksnodeclassification}, C\&S~\cite{huang2020combininglabelpropagationsimple}, and KD~\cite{hinton2015distillingknowledgeneuralnetwork}.
We compare \SAGNAS{} with commonly-used hand-crafted GNN models (i.e., GCN and GAT) in Table~\ref{tbl:trick}.
%
The architecture searched by \SAGNAS{} consistently outperforms GCN and GAT when integrated with all these techniques, demonstrating strong compatibility with other techniques.

\begin{figure*}[t]
\begin{minipage}[t]{1\textwidth} 
\captionsetup[subfloat]{labelfont=scriptsize,textfont=scriptsize}
        \subfloat[CS\label{fig:cs}]{
            \includegraphics[width=0.195\textwidth]{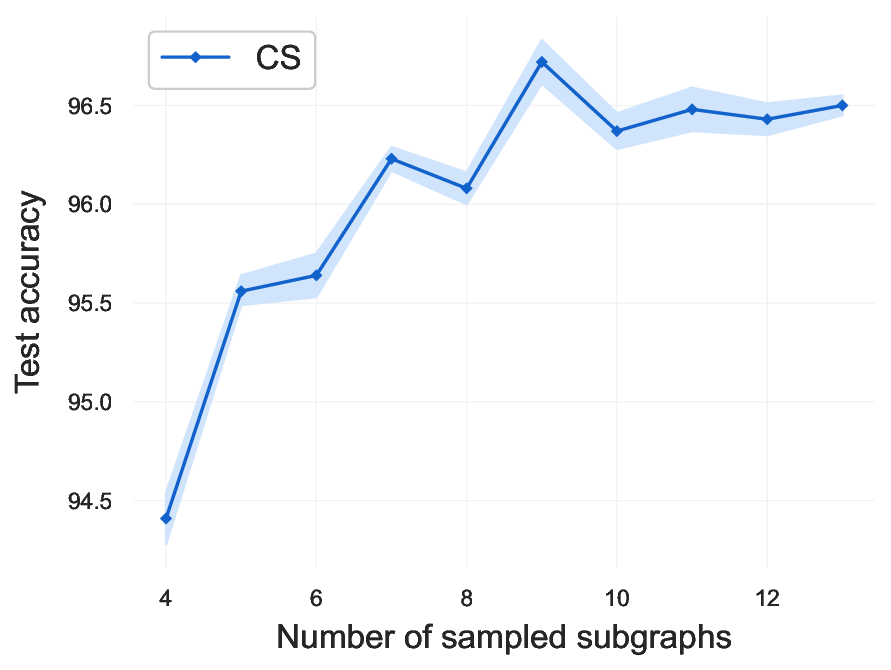}
            }
        \hspace{-1.0em} 
        \subfloat[Physics\label{fig:Physics}]{
            \includegraphics[width=0.195\textwidth]{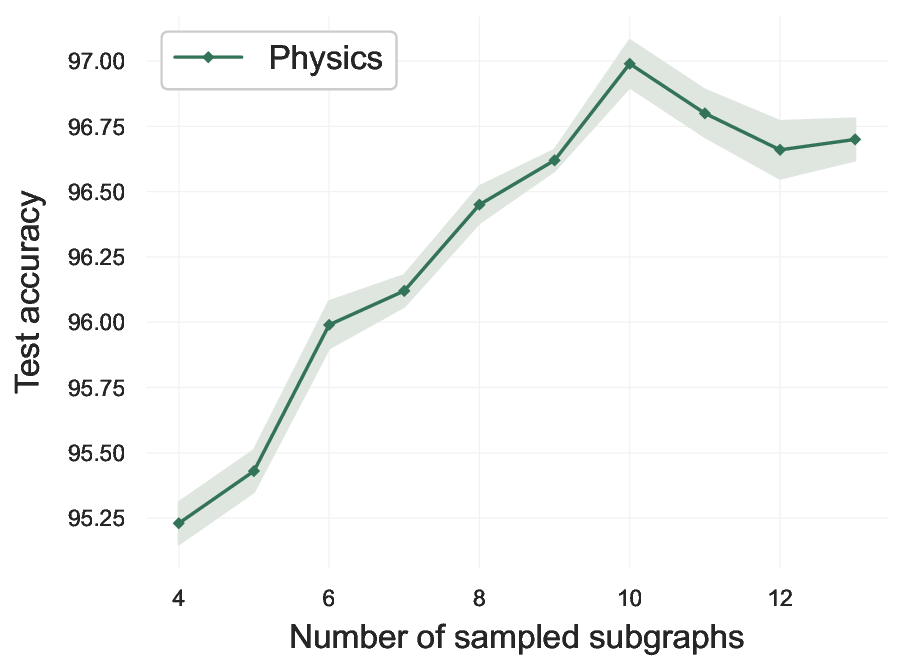}
            
            }
        \hspace{-1.0em}  
        \subfloat[Arxiv\label{fig:arxiv}]{
            \includegraphics[width=0.195\textwidth]{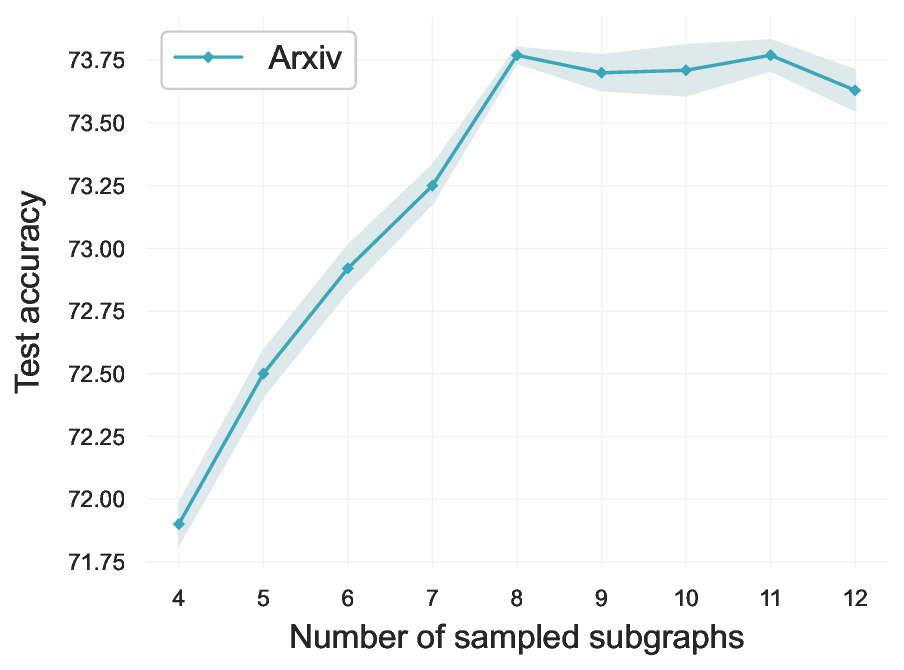}
            
            }
        \hspace{-1.0em}  
        \subfloat[Products\label{fig:products}]{
            \includegraphics[width=0.195\textwidth]{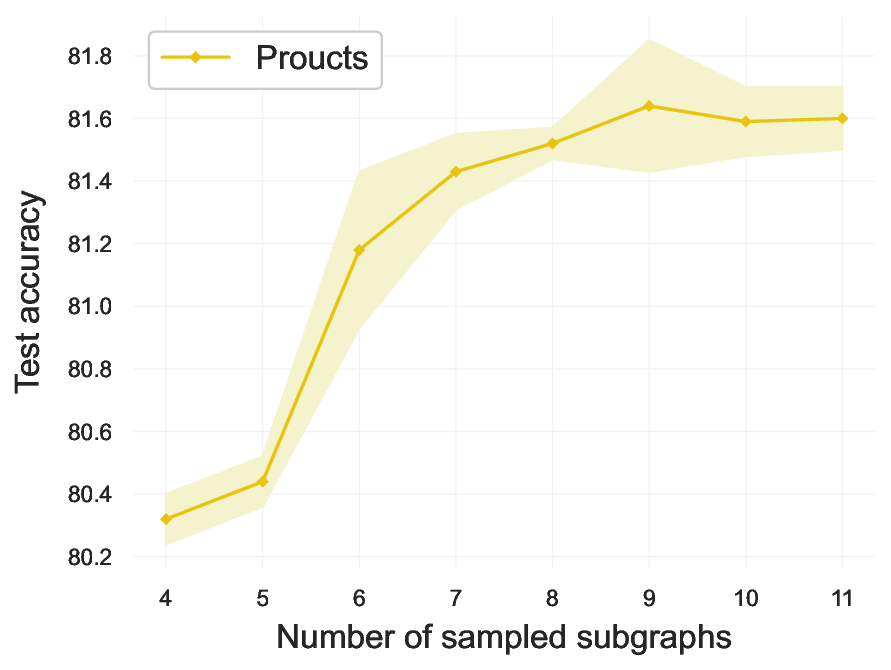}
            
            }
        \hspace{-1.0em} 
        \subfloat[Papers100M\label{fig:papers}]{
            
            \includegraphics[width=0.195\textwidth]{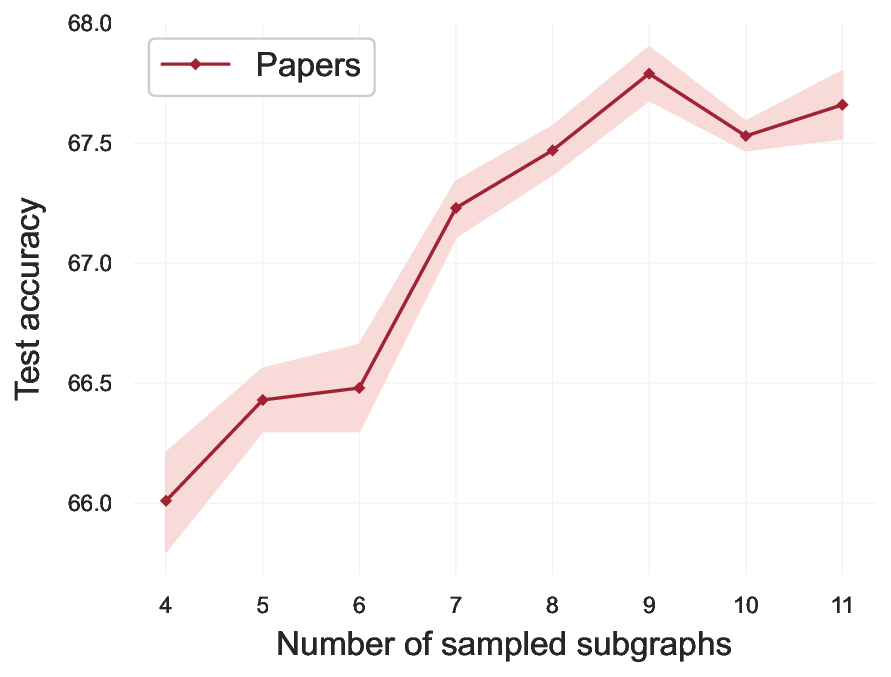}
            
            }
        
        \caption{The influence of the number of sampled subgraphs on the final performance of searched architectures.}
        \label{samplingk}
    \end{minipage}%
\\
\begin{minipage}[t]{1\textwidth}
\captionsetup[subfloat]{labelfont=scriptsize,textfont=scriptsize}
        \subfloat[CS\label{fig:csnodes}]{
        \includegraphics[width=0.195\textwidth]{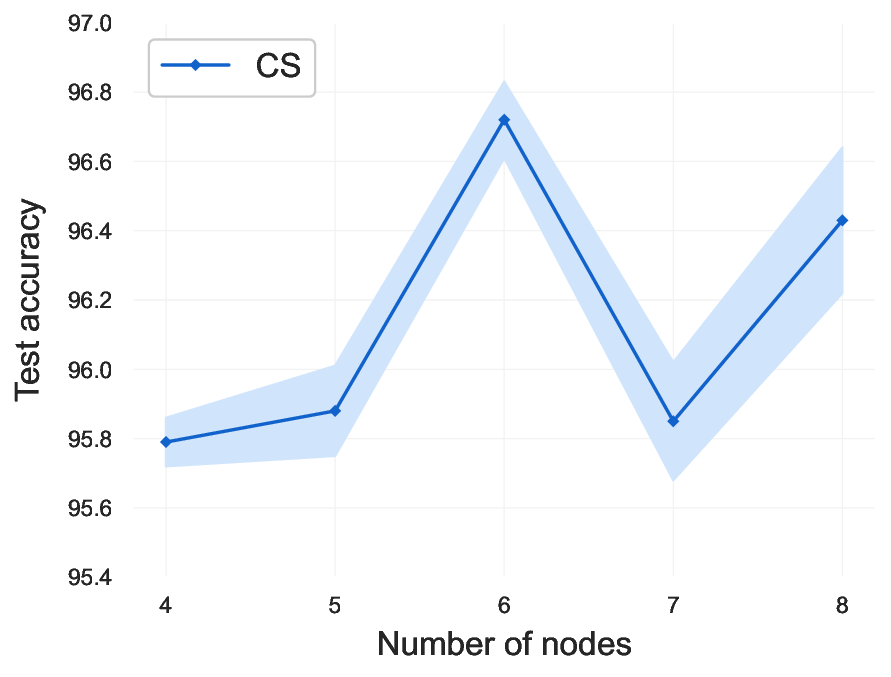}
    }
       \hspace{-1.0em} 
        \subfloat[Physics\label{fig:Physicsnodes}]{
        \includegraphics[width=0.195\textwidth]{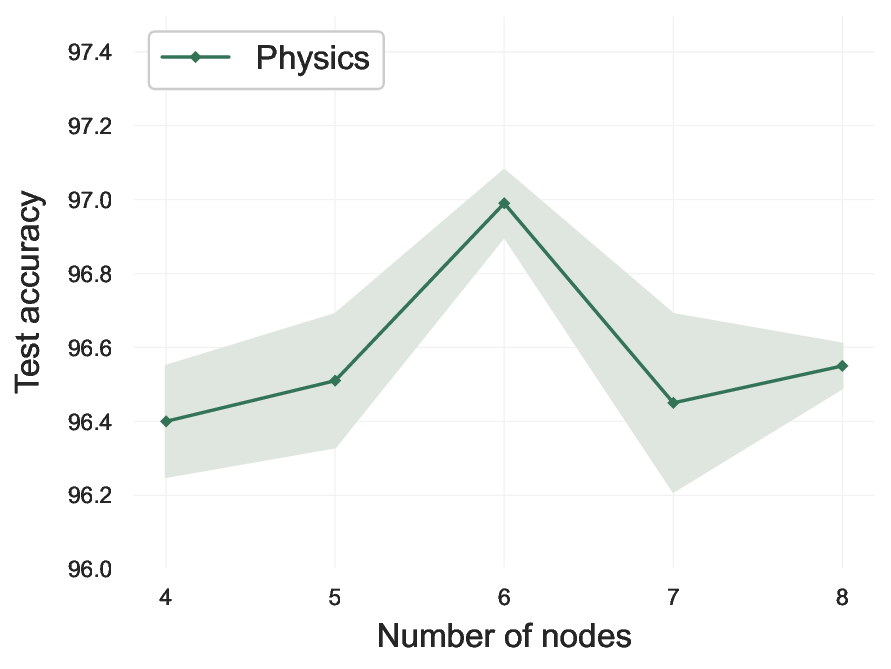}
    }
    \hspace{-1.0em}  
            \subfloat[Arxiv\label{fig:arxivnodes}]{
        \includegraphics[width=0.195\textwidth]{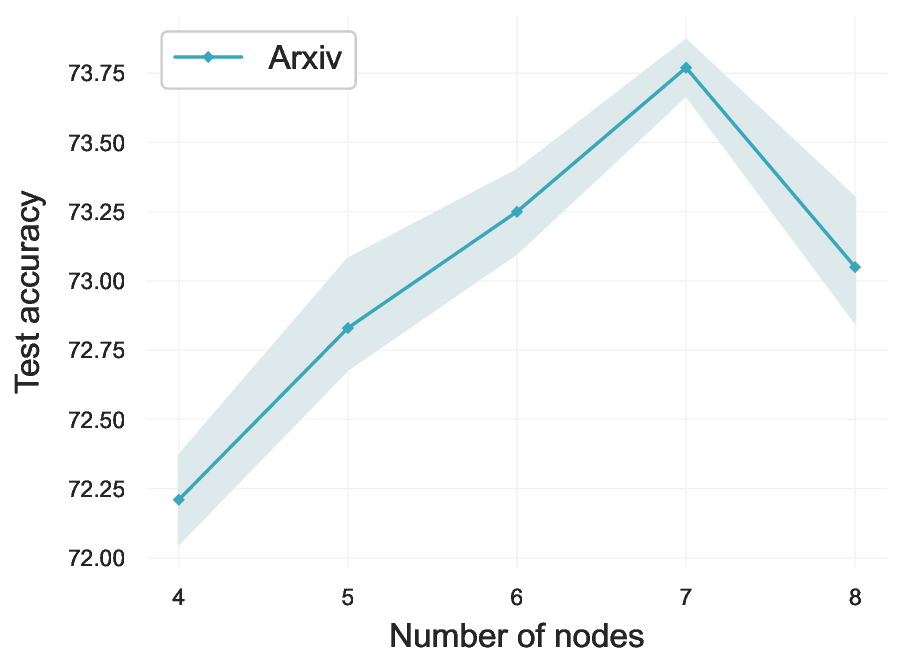}
    }
    \hspace{-1.0em}  
    \subfloat[Products\label{fig:productsnodes}]{
        \includegraphics[width=0.195\textwidth]{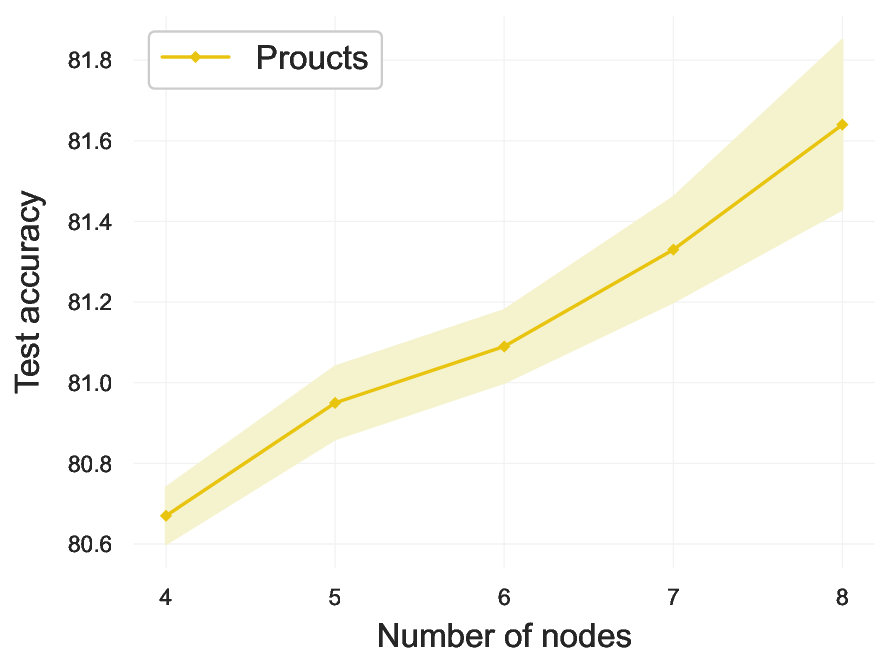}
    }
    \hspace{-1.0em} 
    \subfloat[Papers100M\label{fig:papersnodes}]{
        \includegraphics[width=0.195\textwidth]{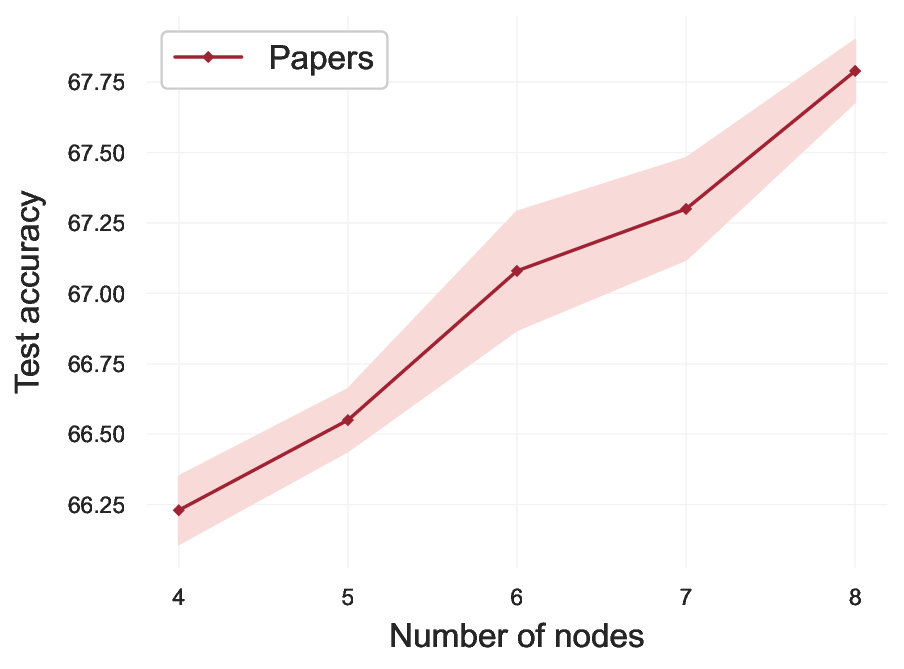}
    }
    \label{fig:papersnodes}
    \caption{The influence of the number of intermediate nodes in the cell on the final performance of searched architectures.}
        \label{nodechoice}
    \end{minipage}%
\\
\begin{minipage}[t]{1\textwidth}
\captionsetup[subfloat]{labelfont=scriptsize,textfont=scriptsize}
    \subfloat[CS\label{fig:cscells}]{
        \includegraphics[width=0.195\textwidth]{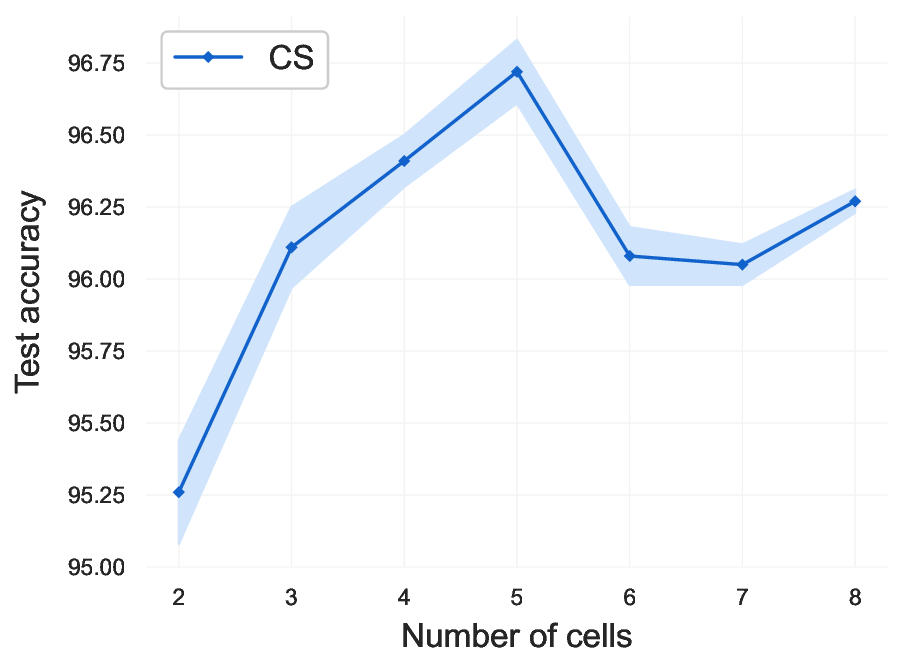}
    }
    \hspace{-1.0em}
    \subfloat[Physics\label{fig:Physicscells}]{
        \includegraphics[width=0.195\textwidth]{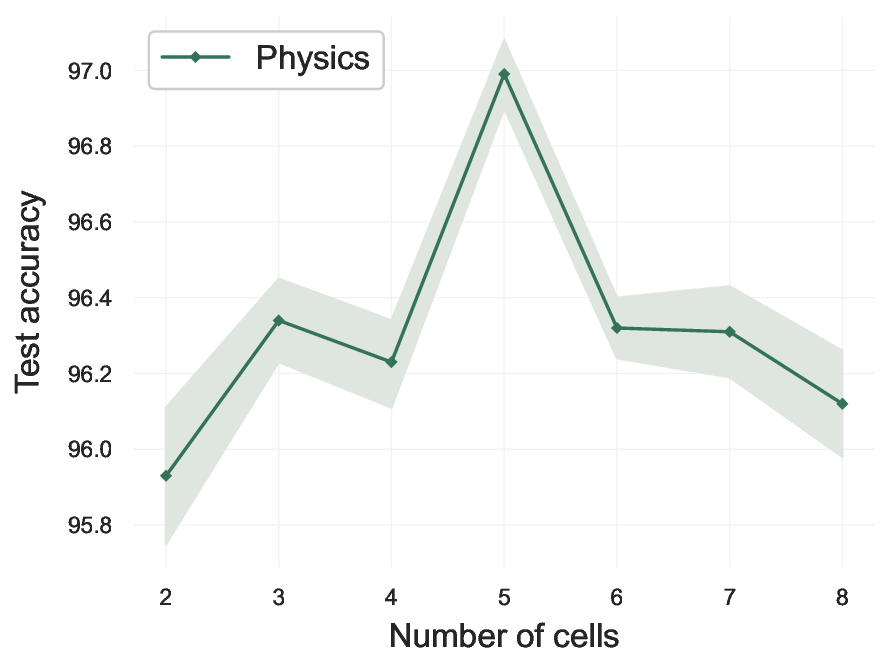}
    }
    \hspace{-1.0em}
    \subfloat[Arxiv\label{fig:arxivcells}]{
        \includegraphics[width=0.195\textwidth]{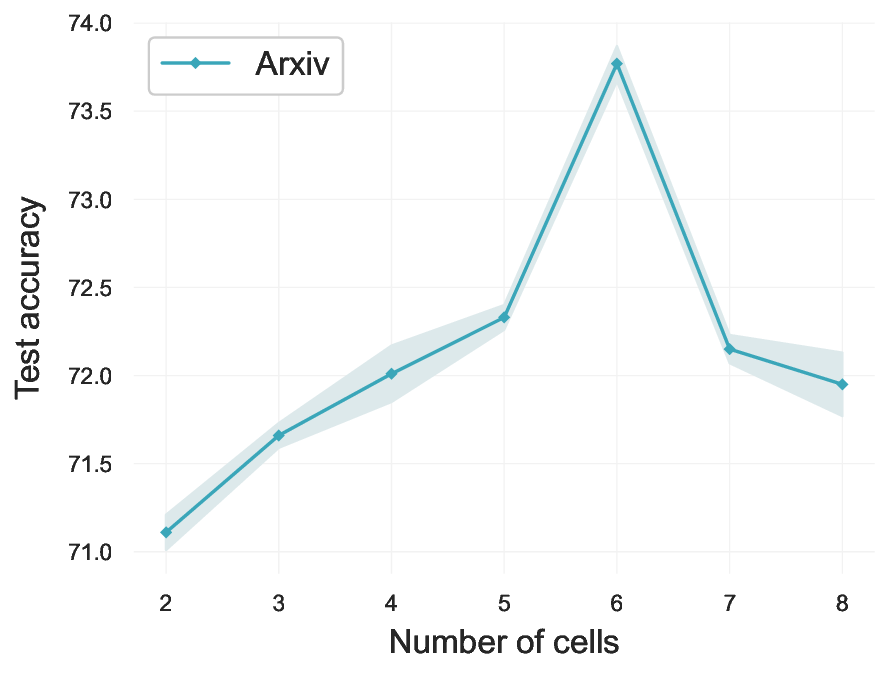}
    }
    \hspace{-1.0em}
    \subfloat[Products\label{fig:productscells}]{
        \includegraphics[width=0.195\textwidth]{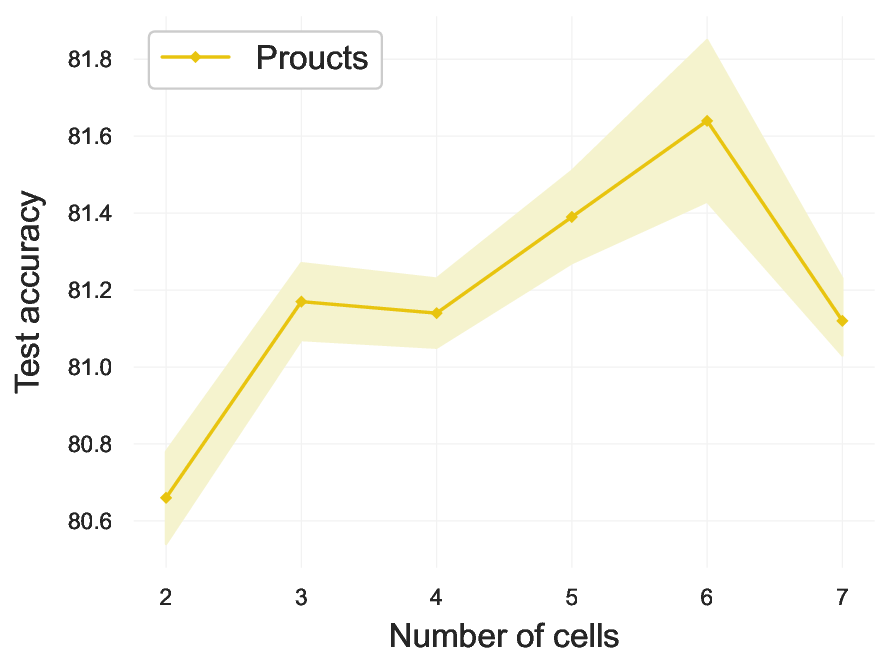}
    }
    \hspace{-1.0em}
    \subfloat[Papers100M\label{fig:paperscells}]{
        \includegraphics[width=0.195\textwidth]{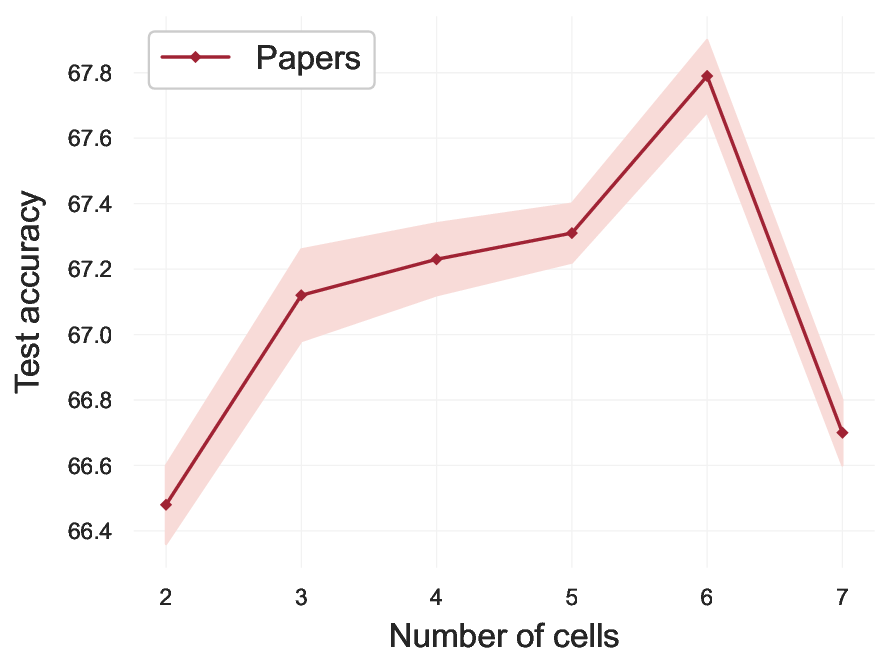}
    }
    \caption{The influence of the number of stacked cells on the final performance of searched architectures.}
    \label{cellchoice}
\end{minipage}%

\end{figure*}

\par
\subsection{Hyperparameter Analysis (RQ4)}
\subsubsection{The number of sampled subgraphs}
we first evaluate the impact of different choices regarding the number of sampled subgraphs in Figure~\ref{samplingk}.
We observe that the performance of the final architecture exhibits an increasing trend as the number of sampled subgraphs (i.e., $K$) rises. 
However, beyond a certain threshold, further increases in $K$ do not result in significant performance improvement. 
If $K$ is too small, the sampled subgraphs are overly random and fail to represent the original large graph adequately. 
Conversely, as the number of subgraphs increases, we are more likely to identify the subgraph that best matches the original graph.

\subsubsection{The number of iterations for architecture expansion}
In Table~\ref{iteration augmentation}, we determine the optimal number of iterations for architecture expansion across various datasets. 
The optimal number of iterations differs depending on the scale of the datasets. 
Larger graphs need more iterations.
By splitting the nodes more times, we can develop more complex architectures as a way to adapt to larger graphs.

\begin{table}[t]
\caption{Performance with different numbers of iterations for architecture expansion. 
}
\scalebox{1}{
\resizebox{\linewidth}{!}{%
\begin{tabular}{@{}lccccc@{}}
\toprule
\multirow{2}{*}{\textbf{Datasets}} & \multicolumn{5}{c}{\textbf{Number of iterations for architecture expansion}}                     \\ \cmidrule(l){2-6} 
                                   & 1     & 2     & 3     & 4              & 5              \\ \midrule
CS                                 & $95.79_{0.07}$ & $95.88_{0.13}$ & \textbf{96.72$_{0.11}$} &$96.85_{0.07}$  & $96.43_{0.21}$          \\
Physics                            & $96.40_{0.15}$ & $96.51_{0.18}$ & \textbf{96.89$_{0.09}$} &$96.45_{0.24}$  & $96.55_{0.06}$          \\
Arxiv                              & $72.21_{0.16}$ & $72.83_{0.25}$ & $72.25_{0.15}$ & \textbf{73.77$_{0.10}$} & $73.05_{0.20}$          \\
Products                           & $80.67_{0.07}$ & $80.75_{0.19}$ & $81.09_{0.09}$ & $81.33_{0.13}$          & \textbf{81.64$_{0.21}$} \\
Papers100M                         & $66.23_{0.12}$ & $66.55_{0.11}$ & $67.08_{0.21}$ & $67.30_{0.18}$          & \textbf{67.79$_{0.11}$} \\ \bottomrule
\end{tabular}
}
}
\label{iteration augmentation}
\vspace{-3ex}
\end{table}

\begin{figure}[t]
\captionsetup[subfloat]{labelfont=scriptsize,textfont=scriptsize}
    \subfloat[Coauthor CS\label{fig:coauthor_cs}]{
        \includegraphics[width=0.45\textwidth]{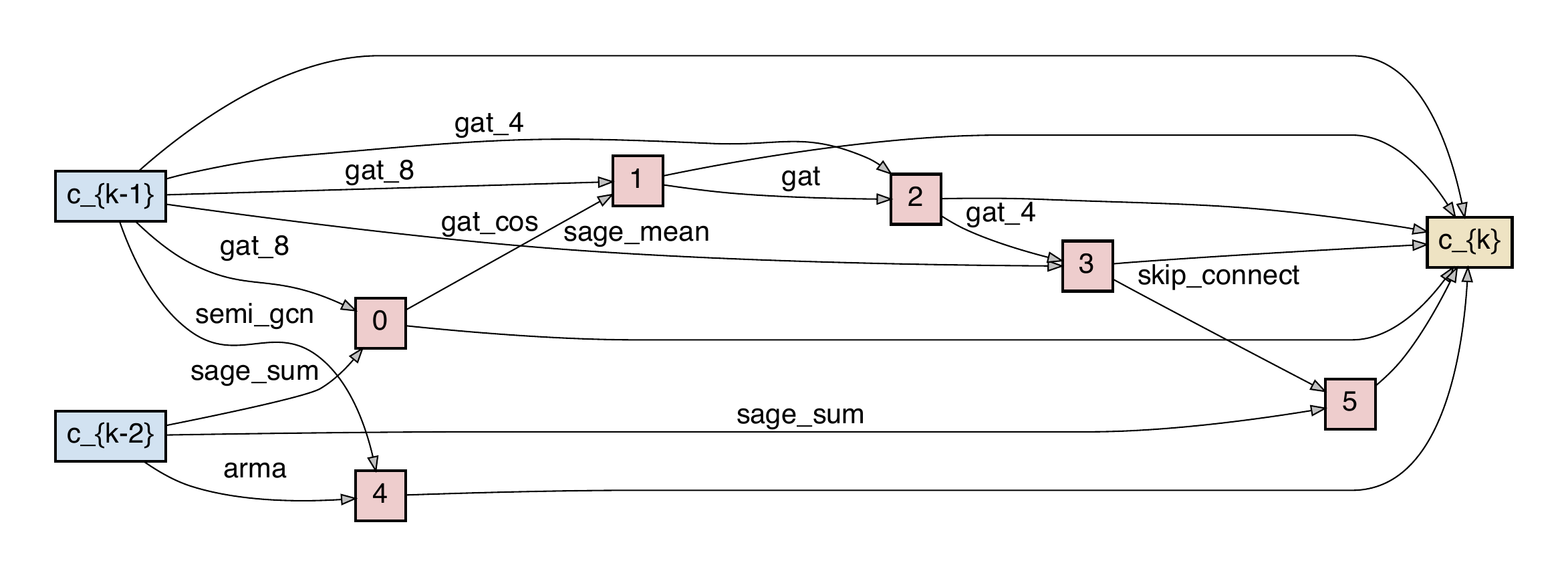}
    }
    \\
    \captionsetup[subfloat]{labelfont=scriptsize,textfont=scriptsize}
    \subfloat[Coauthor Physics\label{fig:coauthor_physics}]{
        \includegraphics[width=0.45\textwidth]{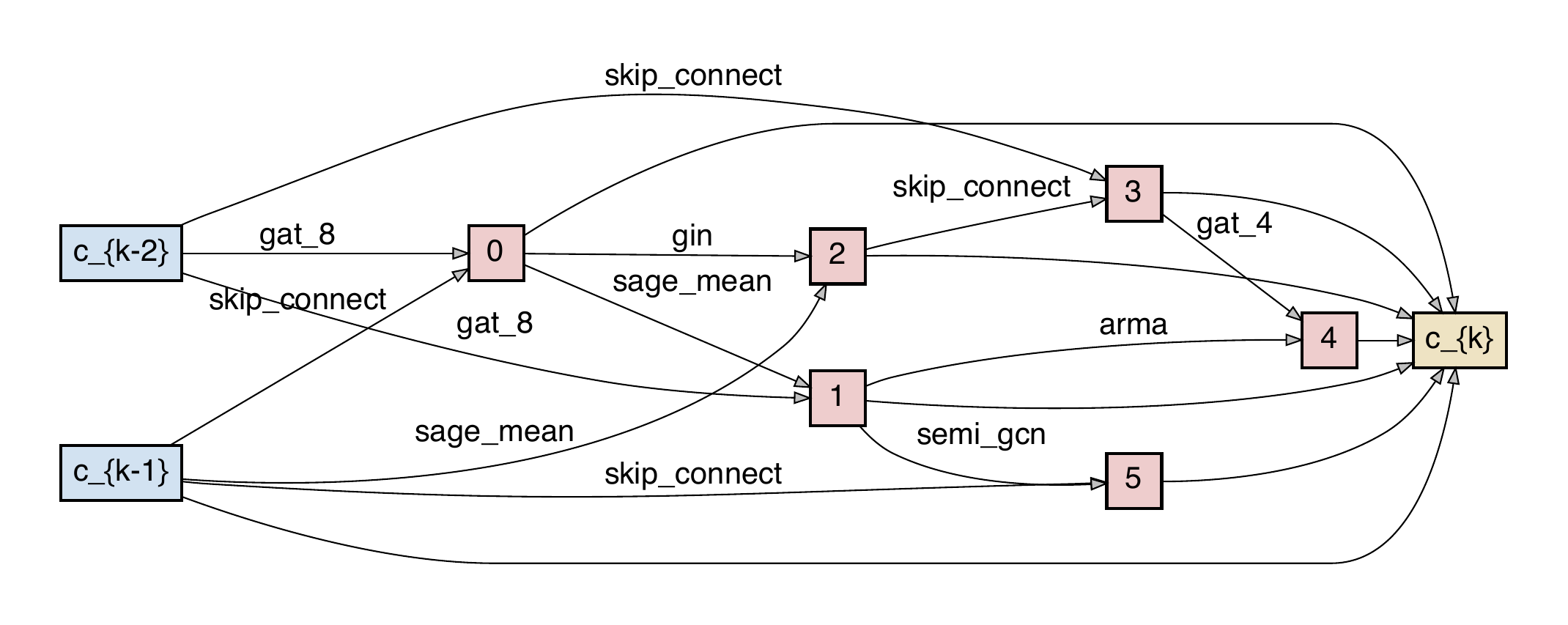}
    }
    \\
    \captionsetup[subfloat]{labelfont=scriptsize,textfont=scriptsize}
    \subfloat[Ogbn-Arxiv\label{fig:arxiv_best}]{
        \includegraphics[width=0.45\textwidth]{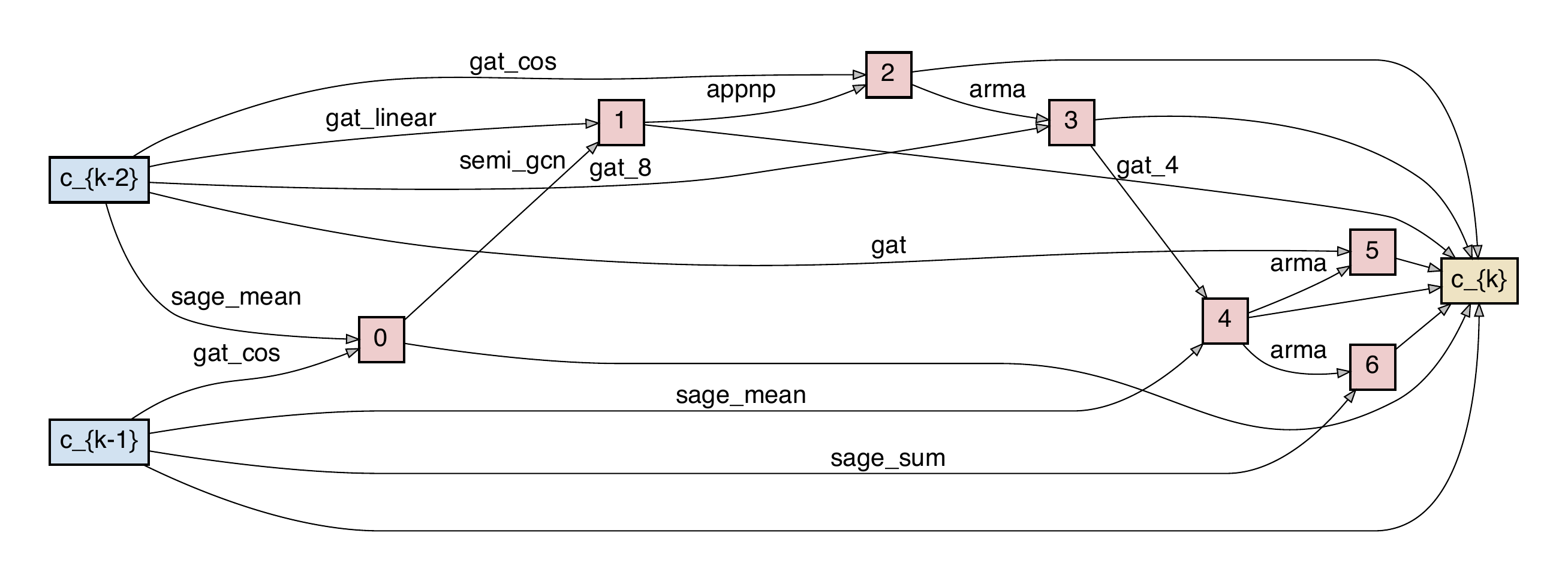}
    }
    \\
    \captionsetup[subfloat]{labelfont=scriptsize,textfont=scriptsize}
    \subfloat[Ogbn-Products\label{fig:products_best}]{
        \includegraphics[width=0.45\textwidth]{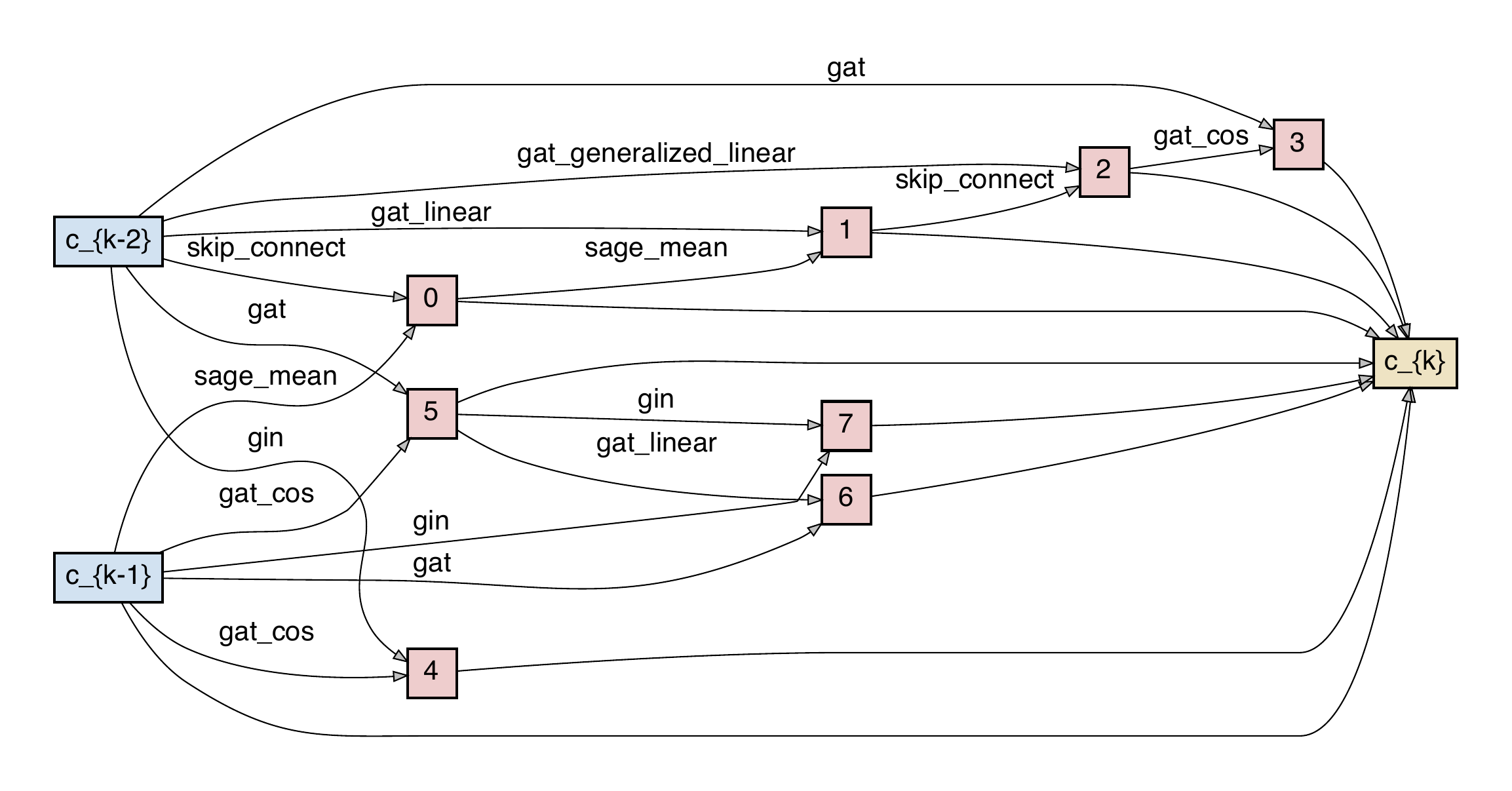}
    }
    \\
    \captionsetup[subfloat]{labelfont=scriptsize,textfont=scriptsize}
    \subfloat[Ogbn-Papers100M\label{fig:papers_best}]{
        \includegraphics[width=0.45\textwidth]{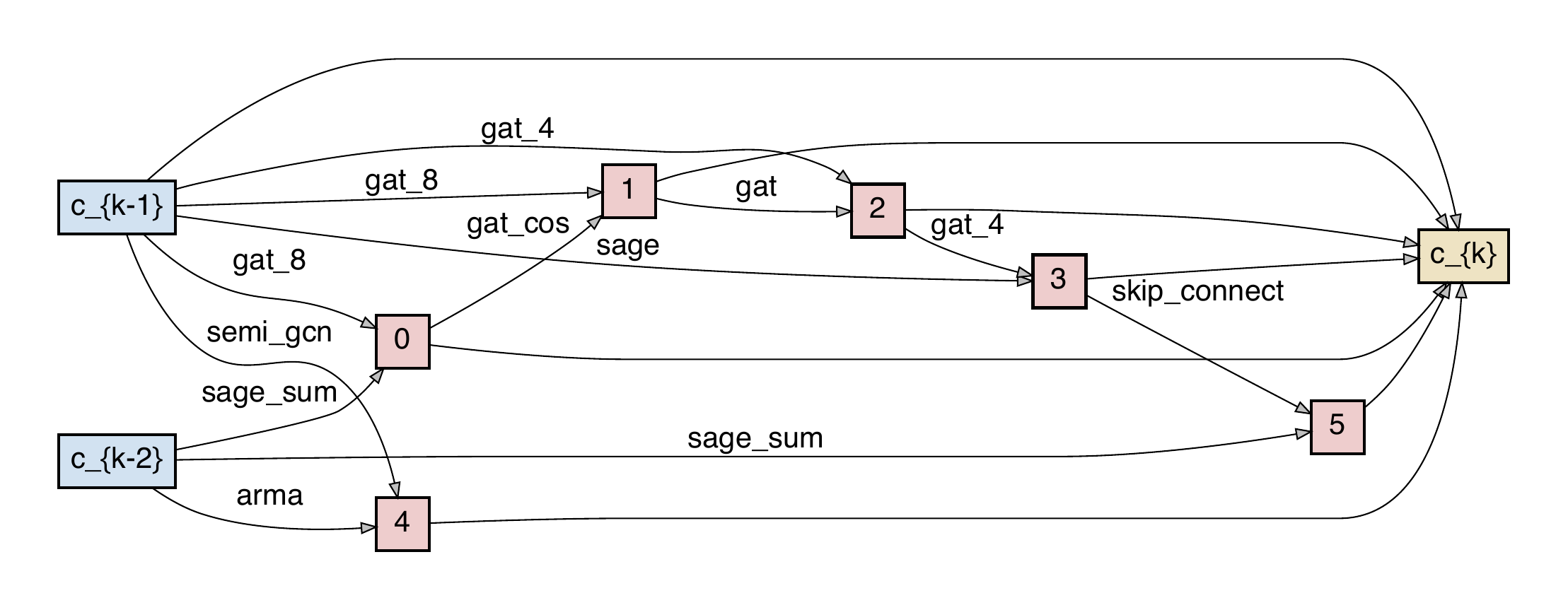}
    }
    \caption{The architectures searched on five large-scale datasets.}
    \label{visarch}
    \vspace{-3ex}
\end{figure}

\subsubsection{The number of intermediate nodes in the cell}
\label{nodestudy}
We further evaluate the influence of different numbers of intermediate nodes in the cell architecture.
As illustrated in Figure~\ref{nodechoice}, each dataset has an optimal configuration that yields the best performance.
However, as the number of intermediate nodes in the cell continues to increase, overall performance begins to decline.
This indicates that the supernet becomes overly complex, leading to issues such as overfitting. 
Furthermore, to handle larger graphs, more intermediate nodes are required to generate more complex cell architectures.

\subsubsection{The number of stacked cells}
We further conduct experiments with different numbers of stacked cells.
The results in Figure~\ref{cellchoice} exhibit a similar trend with that in Figure~\ref{nodechoice}.
The number of stacked cells and the number of nodes within each cell directly influence the complexity of the model.
Larger datasets often necessitate more sophisticated architectures.

\subsection{Visualization (RQ5)}
We visualize the optimal cell architectures searched by \SAGNAS{} in Figure~\ref{visarch}.
All architectures incorporate various types of operations, including parametric operations (e.g., SAGE, GIN, ARMA, and GAT) and non-parametric operations (e.g., skip\_connect). 
Moreover, different datasets correspond to distinct optimal architectures, underscoring the necessity of conducting GNN architecture search.
The proposed \SAGNAS{} can obtain the most suitable
GNN architecture automatically for a specific graph dataset.

\section{Conclusion}
In this paper, we proposed a novel framework based on \textit{seed architecture expansion} for efficient large-scale GNAS. 
The search process is inspired by \emph{cell expansion} in biotechnology and consists of two stages: seed architecture selection and seed architecture expansion.
In the first stage, we proposed a performance ranking consistency-based seed architecture selection method, which employs weighted Kendall $\tau$ coefficient to measure the correlation of performance sequences between each subgraph and the original graph.
The architecture searched on the subgraph that best matches the original large-scale graph is designated as the seed architecture.
In the second stage, we proposed an entropy minimization-based seed architecture expansion method, which collaboratively increases both subgraph size and architecture complexity in each iteration.
%
%
Extensive experimental results on five large-scale graphs demonstrate that the proposed \SAGNAS{} outperforms the SOTA human-crafted GNN architectures and existing GNAS methods in terms of both effectiveness and efficiency.


\section*{Acknowledgment}
This work was supported by the National Natural Science Foundation of China (\#62102177), the Frontier Technology R\&D Program of Jiangsu Province (\#BF2024005), and the Collaborative Innovation Center of Novel Software Technology and Industrialization, Jiangsu, China.



\bibliographystyle{IEEEtran}
\bibliography{main}

\begin{thebibliography}{10}
\providecommand{\url}[1]{#1}
\csname url@samestyle\endcsname
\providecommand{\newblock}{\relax}
\providecommand{\bibinfo}[2]{#2}
\providecommand{\BIBentrySTDinterwordspacing}{\spaceskip=0pt\relax}
\providecommand{\BIBentryALTinterwordstretchfactor}{4}
\providecommand{\BIBentryALTinterwordspacing}{\spaceskip=\fontdimen2\font plus
\BIBentryALTinterwordstretchfactor\fontdimen3\font minus \fontdimen4\font\relax}
\providecommand{\BIBforeignlanguage}[2]{{%
\expandafter\ifx\csname l@#1\endcsname\relax
\typeout{** WARNING: IEEEtran.bst: No hyphenation pattern has been}%
\typeout{** loaded for the language `#1'. Using the pattern for}%
\typeout{** the default language instead.}%
\else
\language=\csname l@#1\endcsname
\fi
#2}}
\providecommand{\BIBdecl}{\relax}
\BIBdecl

\bibitem{hamilton2018representation}
W.~L. Hamilton, R.~Ying, and J.~Leskovec, ``Representation learning on graphs: Methods and applications,'' \emph{{IEEE} Data Eng. Bull.}, vol.~40, no.~3, pp. 52--74, 2017.

\bibitem{kipf2017gcn}
T.~N. Kipf and M.~Welling, ``Semi-supervised classification with graph convolutional networks,'' in \emph{Proceedings of the International Conference on Learning Representations}, 2017.

\bibitem{velivckovic2018gat}
P.~Veli{\v{c}}kovi{\'c}, G.~Cucurull, A.~Casanova, A.~Romero, P.~Lio, and Y.~Bengio, ``Graph attention networks,'' in \emph{International Conference on Learning Representations}, 2018.

\bibitem{hamilton2017sage}
W.~L. Hamilton, R.~Ying, and J.~Leskovec, ``Inductive representation learning on large graphs,'' in \emph{International Conference on Neural Information Processing Systems}, 2017, pp. 1025--1035.

\bibitem{pmlr-v162-wang22am}
X.~Wang and M.~Zhang, ``How powerful are spectral graph neural networks,'' in \emph{Proceedings of the 39th International Conference on Machine Learning}, vol. 162, 2022, pp. 23\,341--23\,362.

\bibitem{graphtraffic}
W.~Jiang and J.~Luo, ``Graph neural network for traffic forecasting: A survey,'' \emph{Expert Systems with Applications}, vol. 207, no.~C, Nov. 2022.

\bibitem{graphcombinatorial}
Q.~Cappart, D.~Ch\'{e}telat, E.~B. Khalil, A.~Lodi, C.~Morris, and P.~Veli\v{c}kovi\'{c}, ``Combinatorial optimization and reasoning with graph neural networks,'' \emph{Journal of Machine Learning Research}, vol.~24, no.~1, 2024.

\bibitem{graphmecular}
J.~G. Rittig, Q.~Gao, M.~Dahmen, A.~Mitsos, and A.~M. Schweidtmann, \emph{Graph Neural Networks for the Prediction of Molecular Structure–Property Relationships}.\hskip 1em plus 0.5em minus 0.4em\relax Royal Society of Chemistry, Dec. 2023, pp. 159--181.

\bibitem{graphrecomm}
S.~Wu, F.~Sun, W.~Zhang, X.~Xie, and B.~Cui, ``Graph neural networks in recommender systems: A survey,'' \emph{ACM Computing Surveys}, vol.~55, no.~5, pp. 1--37, 2022.

\bibitem{gao2022hgnn+}
Y.~Gao, Y.~Feng, S.~Ji, and R.~Ji, ``Hgnn+: General hypergraph neural networks,'' \emph{IEEE Transactions on Pattern Analysis and Machine Intelligence}, vol.~45, no.~3, pp. 3181--3199, 2022.

\bibitem{tiezzi2021deep}
M.~Tiezzi, G.~Marra, S.~Melacci, and M.~Maggini, ``Deep constraint-based propagation in graph neural networks,'' \emph{IEEE Transactions on Pattern Analysis and Machine Intelligence}, vol.~44, no.~2, pp. 727--739, 2021.

\bibitem{zhang2022unsupervised}
R.~Zhang, Y.~Zhang, C.~Lu, and X.~Li, ``Unsupervised graph embedding via adaptive graph learning,'' \emph{IEEE Transactions on Pattern Analysis and Machine Intelligence}, vol.~45, no.~4, pp. 5329--5336, 2022.

\bibitem{GNAS-Survey}
Z.~Zhang, X.~Wang, and W.~Zhu, ``Automated machine learning on graphs: A survey,'' in \emph{Proceedings of the Thirtieth International Joint Conference on Artificial Intelligence}, 2021, pp. 4704--4712.

\bibitem{gao2020graphnas}
Y.~Gao, H.~Yang, P.~Zhang, C.~Zhou, and Y.~Hu, ``Graph neural architecture search,'' in \emph{International Joint Conference on Artificial Intelligence}, 2020, pp. 1403--1409.

\bibitem{gasso}
Y.~Qin, X.~Wang, Z.~Zhang, and W.~Zhu, ``Graph differentiable architecture search with structure learning,'' in \emph{Advances in Neural Information Processing Systems}, 2021.

\bibitem{autognn}
\BIBentryALTinterwordspacing
K.~Zhou, Q.~Song, X.~Huang, and X.~Hu, ``Auto-gnn: Neural architecture search of graph neural networks,'' 2019. [Online]. Available: \url{https://arxiv.org/abs/1909.03184}
\BIBentrySTDinterwordspacing

\bibitem{sane}
H.~ZHAO, Q.~YAO, and W.~TU, ``Search to aggregate neighborhood for graph neural network,'' in \emph{2021 IEEE 37th International Conference on Data Engineering (ICDE)}, 2021, pp. 552--563.

\bibitem{psp}
G.~Zhu, W.~Wang, Z.~Xu, F.~Cheng, M.~Qiu, C.~Yuan, and Y.~Huang, ``Psp: Progressive space pruning for efficient graph neural architecture search,'' in \emph{2022 IEEE 38th International Conference on Data Engineering (ICDE)}, 2022, pp. 2168--2181.

\bibitem{gauss}
C.~Guan, X.~Wang, H.~Chen, Z.~Zhang, and W.~Zhu, ``Large-scale graph neural architecture search,'' in \emph{Proceedings of the 39th International Conference on Machine Learning}, vol. 162, 2022, pp. 7968--7981.

\bibitem{hu2021open}
W.~Hu, M.~Fey, M.~Zitnik, Y.~Dong, H.~Ren, B.~Liu, M.~Catasta, and J.~Leskovec, ``Open graph benchmark: Datasets for machine learning on graphs,'' in \emph{Advances in Neural Information Processing Systems}, vol.~33, 2020, pp. 22\,118--22\,133.

\bibitem{Jia2020}
Z.~Jia, S.~Lin, R.~Ying, J.~You, J.~Leskovec, and A.~Aiken, ``Redundancy-free computation for graph neural networks,'' in \emph{Proceedings of the 26th ACM SIGKDD International Conference on Knowledge Discovery and Data Mining}, 2020, pp. 997--1005.

\bibitem{gcn-cluster}
W.-L. Chiang, X.~Liu, S.~Si, Y.~Li, S.~Bengio, and C.-J. Hsieh, ``Cluster-gcn: An efficient algorithm for training deep and large graph convolutional networks,'' in \emph{Proceedings of the 25th ACM SIGKDD International Conference on Knowledge Discovery \& Data Mining}, 2019, pp. 257--266.

\bibitem{graphsaint}
H.~Zeng, H.~Zhou, A.~Srivastava, R.~Kannan, and V.~Prasanna, ``Graphsaint: Graph sampling based inductive learning method,'' in \emph{Proceedings of the International Conference on Learning Representations}, 2020.

\bibitem{LayerNeighborSampling}
M.~F. Balin and U.~\c{C}ataly\"{u}rek, ``Layer-neighbor sampling --- defusing neighborhood explosion in gnns,'' in \emph{Advances in Neural Information Processing Systems}, vol.~36, 2023, pp. 25\,819--25\,836.

\bibitem{EGAN}
\BIBentryALTinterwordspacing
H.~Zhao, L.~Wei, quanming yao, and Z.~He, ``Efficient graph neural architecture search,'' 2021. [Online]. Available: \url{https://openreview.net/forum?id=IjIzIOkK2D6}
\BIBentrySTDinterwordspacing

\bibitem{nas2019jmlr}
T.~Elsken, J.~H. Metzen, and F.~Hutter, ``Neural architecture search: {A} survey,'' \emph{Journal of Machine Learning Research}, vol.~20, pp. 55:1--55:21, 2019.

\bibitem{he2021automl}
X.~He, K.~Zhao, and X.~Chu, ``Automl: A survey of the state-of-the-art,'' \emph{Knowledge-Based Systems}, vol. 212, p. 106622, 2021.

\bibitem{Ren2021Survey}
P.~Ren, Y.~Xiao, X.~Chang, P.-y. Huang, Z.~Li, X.~Chen, and X.~Wang, ``A comprehensive survey of neural architecture search: Challenges and solutions,'' \emph{ACM Computing Surveys}, vol.~54, no.~4, 2021.

\bibitem{chen2023understanding}
W.~Chen, X.~Gong, J.~Wu, Y.~Wei, H.~Shi, Z.~Yan, Y.~Yang, and Z.~Wang, ``Understanding and accelerating neural architecture search with training-free and theory-grounded metrics,'' \emph{IEEE Transactions on Pattern Analysis and Machine Intelligence}, 2023.

\bibitem{zheng2021migo}
X.~Zheng, R.~Ji, Y.~Chen, Q.~Wang, B.~Zhang, J.~Chen, Q.~Ye, F.~Huang, and Y.~Tian, ``Migo-nas: Towards fast and generalizable neural architecture search,'' \emph{IEEE Transactions on Pattern Analysis and Machine Intelligence}, vol.~43, no.~9, pp. 2936--2952, 2021.

\bibitem{zhang2020you}
X.~Zhang, Z.~Huang, N.~Wang, S.~Xiang, and C.~Pan, ``You only search once: Single shot neural architecture search via direct sparse optimization,'' \emph{IEEE Transactions on Pattern Analysis and Machine Intelligence}, vol.~43, no.~9, pp. 2891--2904, 2020.

\bibitem{chen2023mngnas}
Z.~Chen, G.~Qiu, P.~Li, L.~Zhu, X.~Yang, and B.~Sheng, ``Mngnas: distilling adaptive combination of multiple searched networks for one-shot neural architecture search,'' \emph{IEEE Transactions on Pattern Analysis and Machine Intelligence}, 2023.

\bibitem{wang2021sample}
L.~Wang, S.~Xie, T.~Li, R.~Fonseca, and Y.~Tian, ``Sample-efficient neural architecture search by learning actions for monte carlo tree search,'' \emph{IEEE Transactions on Pattern Analysis and Machine Intelligence}, vol.~44, no.~9, pp. 5503--5515, 2021.

\bibitem{darts}
H.~Liu, K.~Simonyan, and Y.~Yang, ``Darts: Differentiable architecture search,'' in \emph{Proceedings of the International Conference on Learning Representations}, 2019.

\bibitem{nasfpn}
G.~Ghiasi, T.-Y. Lin, and Q.~V. Le, ``Nas-fpn: Learning scalable feature pyramid architecture for object detection,'' in \emph{Proceedings of the IEEE/CVF Conference on Computer Vision and Pattern Recognition (CVPR)}, 2019, pp. 7029--7038.

\bibitem{autodeeplab}
C.~Liu, L.-C. Chen, F.~Schroff, H.~Adam, W.~Hua, A.~L. Yuille, and L.~Fei-Fei, ``Auto-deeplab: Hierarchical neural architecture search for semantic image segmentation,'' in \emph{Proceedings of the IEEE/CVF Conference on Computer Vision and Pattern Recognition (CVPR)}, 2019, pp. 82--92.

\bibitem{tian2021alphagan}
Y.~Tian, L.~Shen, G.~Su, Z.~Li, and W.~Liu, ``Alphagan: Fully differentiable architecture search for generative adversarial networks,'' \emph{IEEE Transactions on Pattern Analysis and Machine Intelligence}, vol.~44, no.~10, pp. 6752--6766, 2021.

\bibitem{nas-ctr}
G.~Zhu, F.~Cheng, D.~Lian, C.~Yuan, and Y.~Huang, ``Nas-ctr: Efficient neural architecture search for click-through rate prediction,'' in \emph{Proceedings of the 45th International ACM SIGIR Conference on Research and Development in Information Retrieval}, 2022, pp. 332--342.

\bibitem{autogsr}
J.~Chen, G.~Zhu, H.~Hou, C.~Yuan, and Y.~Huang, ``Autogsr: Neural architecture search for graph-based session recommendation,'' in \emph{Proceedings of the 45th International ACM SIGIR Conference on Research and Development in Information Retrieval}, 2022, pp. 1694--1704.

\bibitem{zoph2017neural}
B.~Zoph and Q.~V. Le, ``Neural architecture search with reinforcement learning,'' in \emph{Proceedings of the International Conference on Learning Representations}, 2017.

\bibitem{pham2018efficient}
H.~Pham, M.~Guan, B.~Zoph, Q.~Le, and J.~Dean, ``Efficient neural architecture search via parameters sharing,'' in \emph{Proceedings of the International Conference on Machine Learning}, 2018, pp. 4095--4104.

\bibitem{NASNet}
B.~Zoph, V.~Vasudevan, J.~Shlens, and Q.~V. Le, ``Learning transferable architectures for scalable image recognition,'' in \emph{Proceedings of the IEEE/CVF Conference on Computer Vision and Pattern Recognition}, 2018, pp. 8697--8710.

\bibitem{hierarchical}
H.~Liu, K.~Simonyan, O.~Vinyals, C.~Fernando, and K.~Kavukcuoglu, ``Hierarchical representations for efficient architecture search,'' in \emph{Proceedings of the International Conference on Learning Representations}, 2018.

\bibitem{real2019regularized}
E.~Real, A.~Aggarwal, Y.~Huang, and Q.~V. Le, ``Regularized evolution for image classifier architecture search,'' in \emph{Proceedings of the AAAI Conference on Artificial Intelligence}, 2019, pp. 4780--4789.

\bibitem{yao2020efficient}
Q.~Yao, J.~Xu, W.-W. Tu, and Z.~Zhu, ``Efficient neural architecture search via proximal iterations.'' in \emph{AAAI Conference on Artificial Intelligence}, 2020, pp. 6664--6671.

\bibitem{wang2021rethinking}
R.~Wang, M.~Cheng, X.~Chen, X.~Tang, and C.-J. Hsieh, ``Rethinking architecture selection in differentiable nas,'' in \emph{Proceedings of the International Conference on Learning Representation}, 2021.

\bibitem{jiang2024operation}
S.~Jiang, Z.~Ji, G.~Zhu, C.~Yuan, and Y.~Huang, ``Operation-level early stopping for robustifying differentiable nas,'' \emph{Advances in Neural Information Processing Systems}, vol.~36, 2024.

\bibitem{yu2022cyclic}
H.~Yu, H.~Peng, Y.~Huang, J.~Fu, H.~Du, L.~Wang, and H.~Ling, ``Cyclic differentiable architecture search,'' \emph{IEEE transactions on pattern analysis and machine intelligence}, vol.~45, no.~1, pp. 211--228, 2022.

\bibitem{yan2021zeronas}
C.~Yan, X.~Chang, Z.~Li, W.~Guan, Z.~Ge, L.~Zhu, and Q.~Zheng, ``Zeronas: Differentiable generative adversarial networks search for zero-shot learning,'' \emph{IEEE transactions on pattern analysis and machine intelligence}, vol.~44, no.~12, pp. 9733--9740, 2021.

\bibitem{HighPG}
Y.~Ma, H.~Ren, B.~Khailany, H.~Sikka, L.~Luo, K.~Natarajan, and B.~Yu, ``High performance graph convolutionai networks with applications in testability analysis,'' in \emph{Proceedings of the 56th ACM/IEEE Design Automation Conference (DAC)}, 2019, pp. 1--6.

\bibitem{GeneticGNN}
M.~Shi, Y.~Tang, X.~Zhu, Y.~Huang, D.~Wilson, Y.~Zhuang, and J.~Liu, ``Genetic-gnn: Evolutionary architecture search for graph neural networks,'' \emph{Knowledge-Based Systems}, vol. 247, p. 108752, 2022.

\bibitem{autoattend}
C.~Guan, X.~Wang, and W.~Zhu, ``Autoattend: Automated attention representation search,'' in \emph{Proceedings of the 38th International Conference on Machine Learning}, 2021, pp. 3864--3874.

\bibitem{zhao2020simplifying}
\BIBentryALTinterwordspacing
H.~Zhao, L.~Wei, and Q.~Yao, ``Simplifying architecture search for graph neural network,'' 2020. [Online]. Available: \url{https://arxiv.org/abs/2008.11652}
\BIBentrySTDinterwordspacing

\bibitem{dss}
Y.~Li, Z.~Wen, Y.~Wang, and C.~Xu, ``One-shot graph neural architecture search with dynamic search space,'' in \emph{Proceedings of the AAAI Conference on Artificial Intelligence}, 2021.

\bibitem{Meta-GNAS}
Y.~Li, J.~Wu, and T.~Deng, ``Meta-gnas: Meta-reinforcement learning for graph neural architecture search,'' \emph{Engineering Applications of Artificial Intelligence}, vol. 123, p. 106300, 2023.

\bibitem{HGNAS}
Y.~Gao, P.~Zhang, C.~Zhou, H.~Yang, Z.~Li, Y.~Hu, and P.~S. Yu, ``Hgnas++: Efficient architecture search for heterogeneous graph neural networks,'' \emph{IEEE Transactions on Knowledge and Data Engineering}, vol.~35, no.~9, pp. 9448--9461, 2023.

\bibitem{AutoGraph}
Y.~Li and I.~King, \emph{AutoGraph: Automated Graph Neural Network}.\hskip 1em plus 0.5em minus 0.4em\relax Springer International Publishing, 2020, pp. 189--201.

\bibitem{GraphPAS}
J.~Chen, J.~Gao, Y.~Chen, M.~B. Oloulade, T.~Lyu, and Z.~Li, ``Graphpas: Parallel architecture search for graph neural networks,'' in \emph{Proceedings of the 44th International ACM SIGIR Conference on Research and Development in Information Retrieval}, 2021.

\bibitem{sgas}
G.~Li, G.~Qian, I.~C. Delgadillo, M.~Müller, A.~Thabet, and B.~Ghanem, ``Sgas: Sequential greedy architecture search,'' in \emph{2020 IEEE/CVF Conference on Computer Vision and Pattern Recognition (CVPR)}, 2020, pp. 1617--1627.

\bibitem{AutoSTG}
Z.~Pan, S.~Ke, X.~Yang, Y.~Liang, Y.~Yu, J.~Zhang, and Y.~Zheng, ``Autostg: Neural architecture search for predictions of spatio-temporal graph,'' in \emph{Proceedings of the Web Conference}, 2021, p. 1846–1855.

\bibitem{nac}
P.~Xu, L.~Zhang, X.~Liu, J.~Sun, Y.~Zhao, H.~Yang, and B.~Yu, ``Do not train it: A linear neural architecture search of graph neural networks,'' in \emph{Proceedings of the 40th International Conference on Machine Learning}, 2023, pp. 38\,826--38\,847.

\bibitem{wang2022automated}
\BIBentryALTinterwordspacing
X.~Wang, Z.~Zhang, H.~Li, and W.~Zhu, ``Automated graph machine learning: Approaches, libraries, benchmarks and directions,'' 2024. [Online]. Available: \url{https://arxiv.org/abs/2201.01288}
\BIBentrySTDinterwordspacing

\bibitem{pasca}
W.~Zhang, Y.~Shen, Z.~Lin, Y.~Li, X.~Li, W.~Ouyang, Y.~Tao, Z.~Yang, and B.~Cui, ``Pasca: A graph neural architecture search system under the scalable paradigm,'' in \emph{Proceedings of the ACM Web Conference}, 2022.

\bibitem{sasaki2023efficient}
\BIBentryALTinterwordspacing
Y.~Sasaki, ``Efficient and explainable graph neural architecture search via monte-carlo tree search,'' 2023. [Online]. Available: \url{https://arxiv.org/abs/2308.15734}
\BIBentrySTDinterwordspacing

\bibitem{graphpnas}
M.~Li, J.~Y. Liu, L.~Sigal, and R.~Liao, ``Graph{PNAS}: Learning probabilistic graph generators for neural architecture search,'' \emph{Transactions on Machine Learning Research}, 2023.

\bibitem{bianchi2021graph}
F.~M. Bianchi, D.~Grattarola, L.~Livi, and C.~Alippi, ``Graph neural networks with convolutional arma filters,'' \emph{IEEE Transactions on Pattern Analysis and Machine Intelligence}, vol.~44, no.~7, pp. 3496--3507, 2022.

\bibitem{klicpera2018predict}
J.~Klicpera, A.~Bojchevski, and S.~G{\"u}nnemann, ``Predict then propagate: Graph neural networks meet personalized pagerank,'' in \emph{Proceedings of the International Conference on Learning Representations}, 2019.

\bibitem{Puka2011}
L.~Puka, \emph{Kendall's Tau}.\hskip 1em plus 0.5em minus 0.4em\relax Berlin, Heidelberg: Springer Berlin Heidelberg, 2011, pp. 713--715.

\bibitem{SHIEH199817}
G.~S. Shieh, ``A weighted kendall's tau statistic,'' \emph{Statistics \& Probability Letters}, vol.~39, no.~1, pp. 17--24, 1998.

\bibitem{10027699}
Y.~Zhao, S.~Zhang, and L.~Akoglu, ``Toward unsupervised outlier model selection,'' in \emph{2022 IEEE International Conference on Data Mining (ICDM)}, 2022, pp. 773--782.

\bibitem{Wang2020MicrosoftAG}
K.~Wang, Z.~Shen, C.~Huang, C.-H. Wu, Y.~Dong, and A.~Kanakia, ``Microsoft academic graph: When experts are not enough,'' \emph{Quantitative Science Studies}, vol.~1, pp. 396--413, 2020.

\bibitem{AutoGNAS}
J.~Chen, J.~Gao, Y.~Chen, B.~M. Oloulade, T.~Lyu, and Z.~Li, ``Auto-gnas: A parallel graph neural architecture search framework,'' \emph{IEEE Transactions on Parallel and Distributed Systems}, vol.~33, no.~11, pp. 3117--3128, 2022.

\bibitem{node2vector}
A.~Grover and J.~Leskovec, ``node2vec: Scalable feature learning for networks,'' in \emph{Proceedings of the ACM SIGKDD International Conference on Knowledge Discovery and Data Mining}, 2016, pp. 855--864.

\bibitem{wang2021bagtricksnodeclassification}
\BIBentryALTinterwordspacing
Y.~Wang, J.~Jin, W.~Zhang, Y.~Yu, Z.~Zhang, and D.~Wipf, ``Bag of tricks for node classification with graph neural networks,'' 2021. [Online]. Available: \url{https://arxiv.org/abs/2103.13355}
\BIBentrySTDinterwordspacing

\bibitem{huang2020combininglabelpropagationsimple}
\BIBentryALTinterwordspacing
Q.~Huang, H.~He, A.~Singh, S.-N. Lim, and A.~R. Benson, ``Combining label propagation and simple models out-performs graph neural networks,'' 2020. [Online]. Available: \url{https://arxiv.org/abs/2010.13993}
\BIBentrySTDinterwordspacing

\bibitem{hinton2015distillingknowledgeneuralnetwork}
\BIBentryALTinterwordspacing
G.~Hinton, O.~Vinyals, and J.~Dean, ``Distilling the knowledge in a neural network,'' 2015. [Online]. Available: \url{https://arxiv.org/abs/1503.02531}
\BIBentrySTDinterwordspacing

\end{thebibliography}

\begin{IEEEbiography}[{\includegraphics[width=1in,height=1.25in,clip,keepaspectratio]{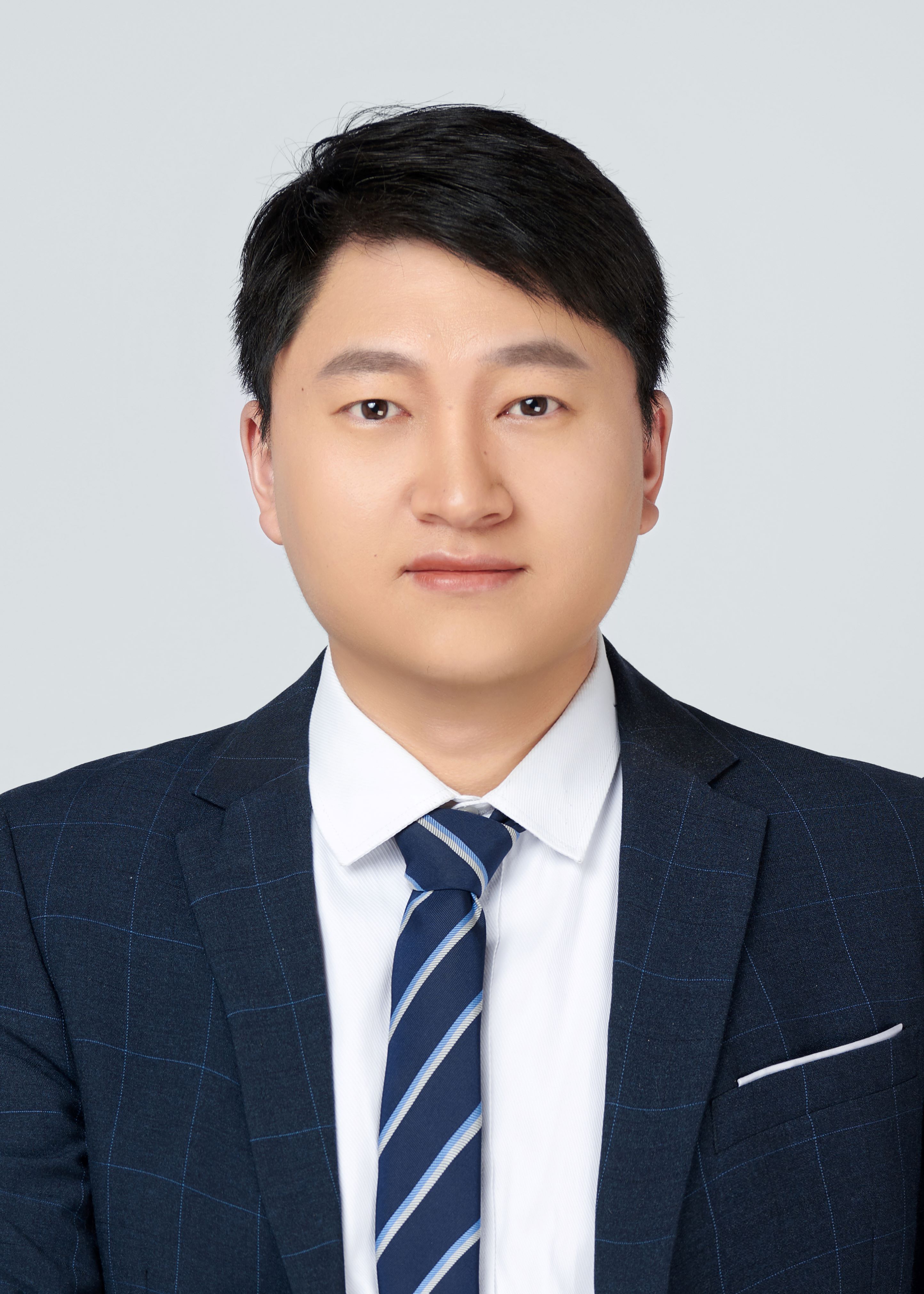}}]{Guanghui Zhu} (Member, IEEE)  is currently an associate researcher in the School of Computer Science, and State Key Laboratory for Novel Software Technology, Nanjing University, China. He received his Ph.D. degrees in computer science and technology from Nanjing University. His main research interests include big data intelligent analysis, graph machine learning, and automated machine learning.
\end{IEEEbiography}
\begin{IEEEbiography}[{\includegraphics[width=1in,height=1.25in,clip,keepaspectratio]{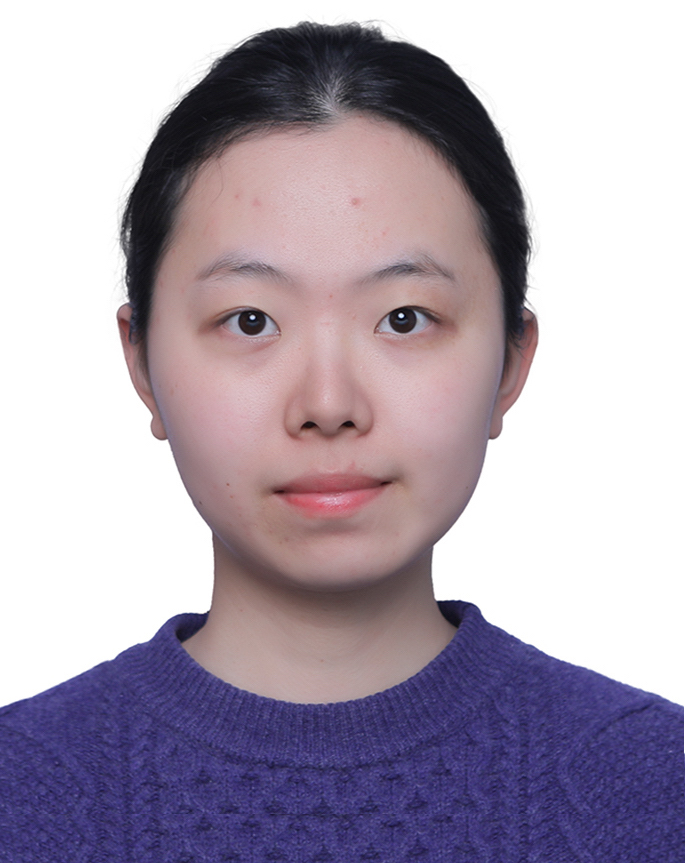}}]{Zipeng Ji} is a postgraduate student in the School of Computer Science, Nanjing University, China. He received his BS degree from Central South University, China. His research interests include graph neural network and automated machine learning.
\end{IEEEbiography}
\begin{IEEEbiography}[{\includegraphics[width=1in,height=1.25in,clip,keepaspectratio]{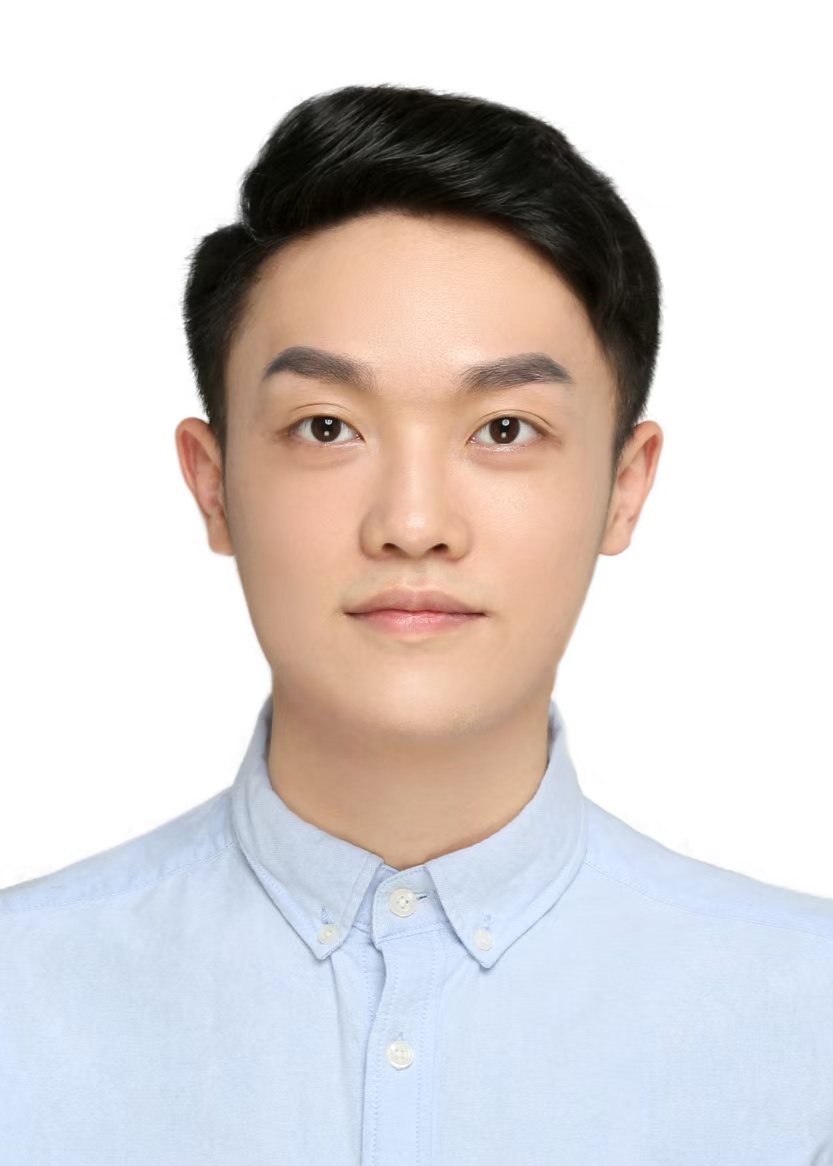}}]{Jingyan Chen} is a postgraduate student in the School of Computer Science, Nanjing University, China. He received his BS degree from South China University of Technology, China. His research interests include graph neural network and graph anomaly detection.
\end{IEEEbiography}
\begin{IEEEbiography}[{\includegraphics[width=1in,height=1.25in,clip,keepaspectratio]{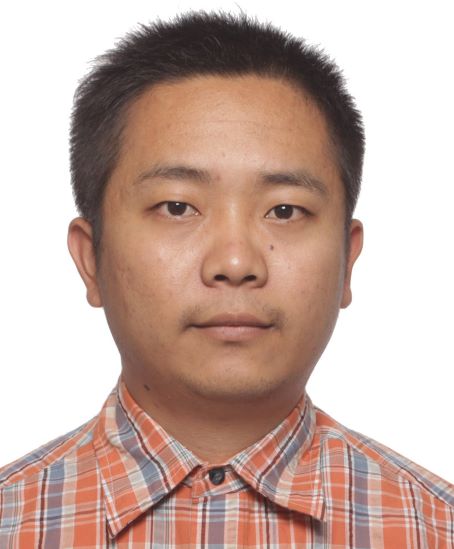}}]{Limin Wang} (Member, IEEE) is currently a professor in the School of Computer Science, Nanjing University, China. He received the B.Sc. degree from Nanjing University in 2011, and the Ph.D. degree from the Chinese University of Hong Kong, Hong Kong, in 2015. From 2015 to 2018, he was a Post-Doctoral Researcher with the Computer Vision Laboratory, ETH Zurich. His research interests include computer vision and deep learning. He has served as an Area Chair for NeurIPS, CVPR, ICCV, and is on the Editorial Board of IJCV and PR.
\end{IEEEbiography}
\begin{IEEEbiography}[{\includegraphics[width=1in,height=1.25in,clip,keepaspectratio]{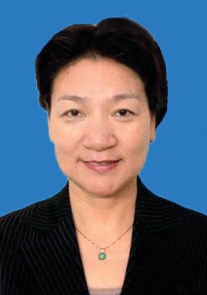}}]{Chunfeng Yuan}
is currently a professor in the School of Computer Science, and State Key Laboratory for Novel Software Technology, Nanjing University, China. She received her bachelor and master degrees in computer science and technology from Nanjing University. Her main research interests include computer architecture, parallel and distributed computing.
\end{IEEEbiography}
\begin{IEEEbiography}[{\includegraphics[width=1in,height=1.25in,clip,keepaspectratio]{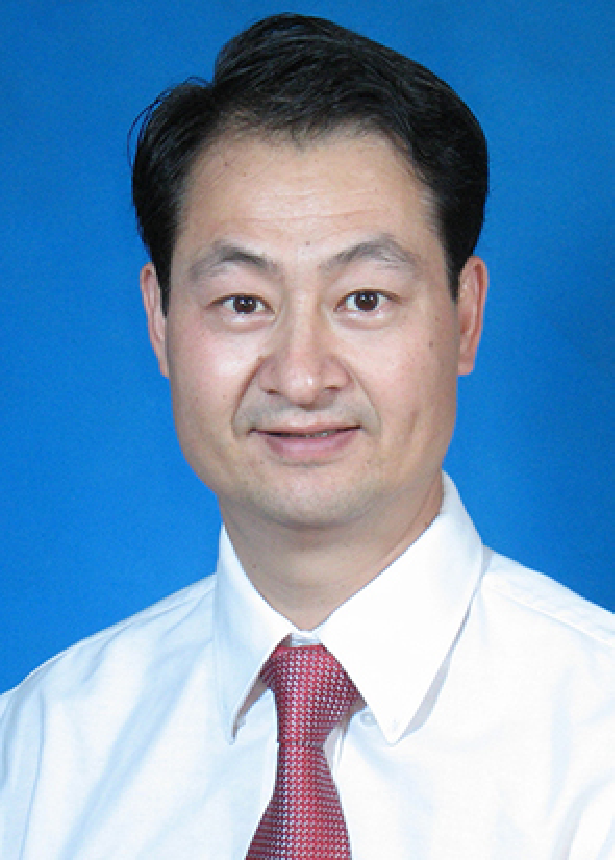}}]{Yihua Huang}
is currently a professor in the School of Computer Science, and State Key Laboratory for Novel Software Technology, Nanjing University, China. He received his bachelor, master and Ph.D. degrees in computer science and technology from Nanjing University. His research interests include parallel and distributed computing, big data parallel processing, big data machine learning algorithm and system.
\end{IEEEbiography}

\vfill

\end{document}